\documentclass{article}

\usepackage[preprint, nonatbib]{neurips}
\makeatletter
\renewcommand{\@noticestring}{}
\makeatother

\usepackage[numbers,sort]{natbib}

\usepackage[export]{adjustbox}
\usepackage[ruled]{algorithm2e}
\usepackage[inline, shortlabels]{enumitem}
\usepackage[T1]{fontenc}
\usepackage{marvosym}        
\usepackage{fontawesome5}    
\usepackage{hyperref}
\usepackage{xspace}
\usepackage{microtype}
\usepackage{pifont}
\usepackage{xcolor}
\usepackage{xurl}
\usepackage{graphicx}
\usepackage{booktabs}
\usepackage{tabularray}
\usepackage{listings}
\usepackage{amsmath, amsfonts}
\usepackage{nicefrac}

\UseTblrLibrary{booktabs}

\makeatletter
\newcommand{\ssymbol}[1]{\@fnsymbol{#1}}
\newcommand{\romanNumeral}[1]{\expandafter\@slowromancap\romannumeral #1@}
\makeatother

\usepackage{subcaption}
\usepackage{tikz}
\usetikzlibrary{arrows.meta, positioning, shapes.geometric, calc, fit, backgrounds}
\usepackage{wrapfig}
\usepackage{multirow}
\usepackage{graphicx}
\usepackage{xcolor}
\usepackage{hyperref}       
\usepackage{booktabs}       
\usepackage{amsfonts}       
\usepackage{nicefrac}       
\usepackage{microtype}      
\usepackage{xcolor}         
\usepackage{amsmath}
\usepackage{amssymb}    
\usepackage{cleveref}   
\usepackage{xspace}     
\DeclareUnicodeCharacter{266B}{\ensuremath{\flat\flat}}  

\newcommand{\cognifold}{\textsc{CogniFold}}

\newcommand{\titleicon}{%
  \hspace{-2.5em}
  \raisebox{-0.48em}{\includegraphics[height=1.95em]{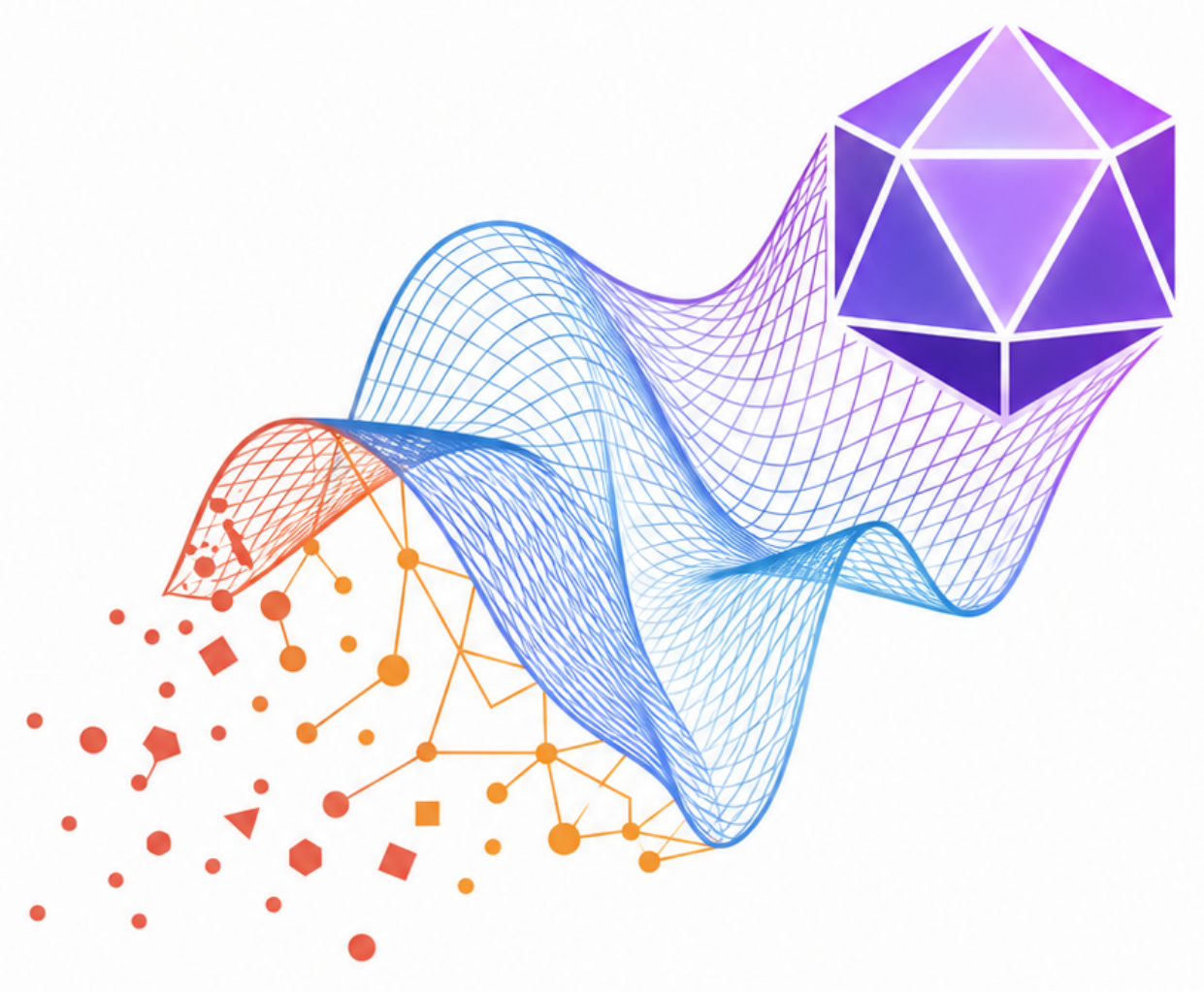}}%
  \hspace{0.5em}
}

\title{
  \vspace{-0.65em}
  \texorpdfstring{\titleicon \cognifold{}:}{\cognifold{}:} Always-On Proactive \\Memory via Cognitive Folding
}

\makeatletter
\renewcommand{\@maketitle}{%
  \vbox{%
    \hsize\textwidth\linewidth\hsize
    \vskip 0.1in
    \@toptitlebar
    \centering
    {\LARGE\bf \@title\par}
    \@bottomtitlebar
    \vspace{0.7em}\par
    {\normalsize
      \begin{tabular}{c}
        \textbf{Suli Wang}\textsuperscript{1,*} \quad
        \textbf{Yiqun Duan}\textsuperscript{1,*,\,\Letter} \quad
        \textbf{Yu Deng}\textsuperscript{1,*} \quad
        \textbf{Rundong Zhao}\textsuperscript{1} \\[0.12cm]
        \textbf{Dai Shi}\textsuperscript{1,2} \quad
        \textbf{Minghua Deng}\textsuperscript{1} \quad
        \textbf{Chen Chen}\textsuperscript{3} \quad
        \textbf{Yiqi Wang}\textsuperscript{4} \quad
        \textbf{Xinliang Zhou}\textsuperscript{1,5} \\[0.3cm]
        \textit{\textsuperscript{1}OpenUploading \quad
                \textsuperscript{2}University of Cambridge \quad
                \textsuperscript{3}NVIDIA \quad
                \textsuperscript{4}Griffith University \quad
                \textsuperscript{5}Stanford University
        }
      \end{tabular}\par
    }
    \vspace{0.2em}\par
    {\normalsize
      \href{https://github.com/OpenUploading/CogniFold}{\faGithub\ \textbf{GitHub}}
      \hspace{2em}
      \href{https://huggingface.co/OpenUploading}{\raisebox{-0.1em}{\includegraphics[height=1em]{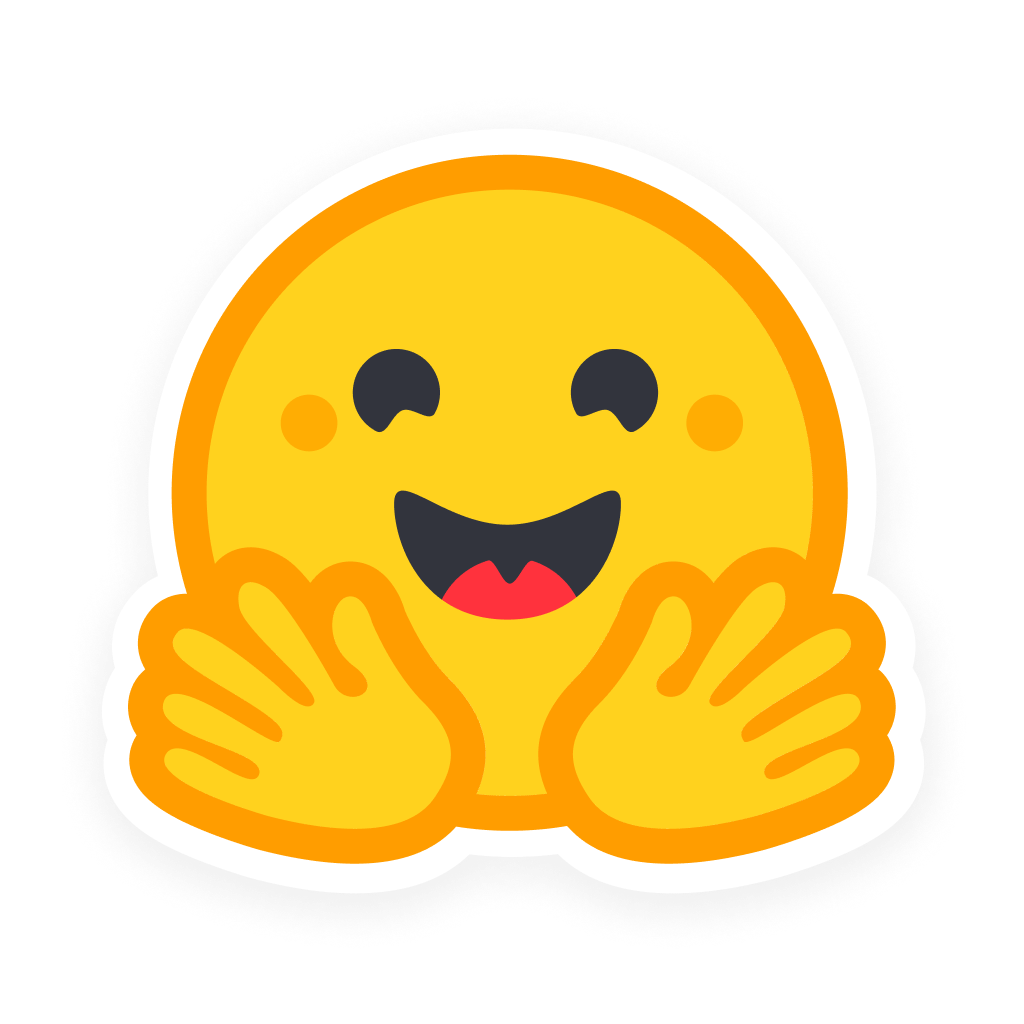}}\ \textbf{HuggingFace}}\par
    }
    \vskip 0.3in \@minus 0.1in
  }
}
\makeatother

\author{}

\begin{document}

\maketitle

\begingroup
\renewcommand\thefootnote{}\footnotetext{$^{*}$Equal contribution. \quad \Letter~Corresponding author: \texttt{duanyiquncc@gmail.com}}
\endgroup

\begin{abstract}
Existing agent memory remains predominantly reactive and retrieval-based, lacking the capacity to autonomously organize experience into persistent cognitive structure. Toward genuinely autonomous agents, we introduce CogniFold, a brain-inspired "always-on" agent memory designed for the next generation of proactive assistants. CogniFold continuously folds fragmented event streams into self-emerging cognitive structures, bootstrapping progressively higher-level cognition from incoming events and accumulated knowledge.
We ground this by extending Complementary Learning Systems (CLS) theory from two layers (hippocampus, neocortex) to three, adding a prefrontal intent layer. Emulating the prefrontal cortex as the locus of intentional control and decision-making, CogniFold achieves this through graph-topology self-organization: cognitive structures proactively assemble under the stream, merge when semantically similar, decay when stale, relink through associative recall, and surface intents when concept-cluster density crosses a threshold.
We evaluate structural formation using CogEval-Bench, demonstrating that CogniFold uniquely produces memory structures that match cognitive expectations and concept emergence. Furthermore, across eight downstream benchmarks---two probing long-term conversational memory (LoCoMo, LongMemEval) and six spanning other cognitive domains---we validate that CogniFold simultaneously performs robustly on conventional memory tasks.
Our code is available at \url{https://github.com/OpenUploading/CogniFold}.
\end{abstract}

\section{Introduction}
\label{sec:intro}

Memory-Augmented Agents~\citep{packer2023memgpt, sumers2023cognitive} have empowered Large Language Models (LLMs) to transcend finite context constraints, enabling long-horizon reasoning~\citep{shinn2023reflexion}, context-grounded personalization~\citep{chhikara2025mem0}, and experience-driven continual learning~\citep{majumder2023clin}. However, as agents evolve from on-demand systems into always-on assistants, their input paradigm shifts from bounded, prompt-driven inputs to continuously arriving, fragmented event streams~\citep{zacks2020event, kurby2008segmentation}. This raises an increasing demand for \textbf{proactive} behaviour: an assistant that self-organizes structure and surfaces intents before the user issues a query. 

Yet, existing memory architectures share a common limit: their topology is fixed once formed. Whether leveraging static knowledge graphs~\citep{gutierrez2024hipporag, gutierrez2025rag}, text-level rewrites~\citep{chhikara2025mem0}, hybrid decoupling~\citep{jiang2026magma}, or temporal tracking~\citep{rasmussen2025zep}, memory remains a graph-as-product---a finished artifact to retrieve from, never a substrate that metabolises under the stream. Consequently, agents are forced to graft proactivity on top as application-layer machinery, such as scheduled triggers, planning loops~\citep{wang2023voyager, yang2024swe}, or periodic reflection~\citep{shinn2023reflexion, xu2025mem}. This separation creates a structural ceiling: intents can only arise from sources the application layer was explicitly designed to handle. We argue that proactivity must instead be a property of the memory substrate---intents should emerge from the topology accumulating the conditions for them.

\begin{figure}[t]
\centering
\includegraphics[width=0.95\textwidth]{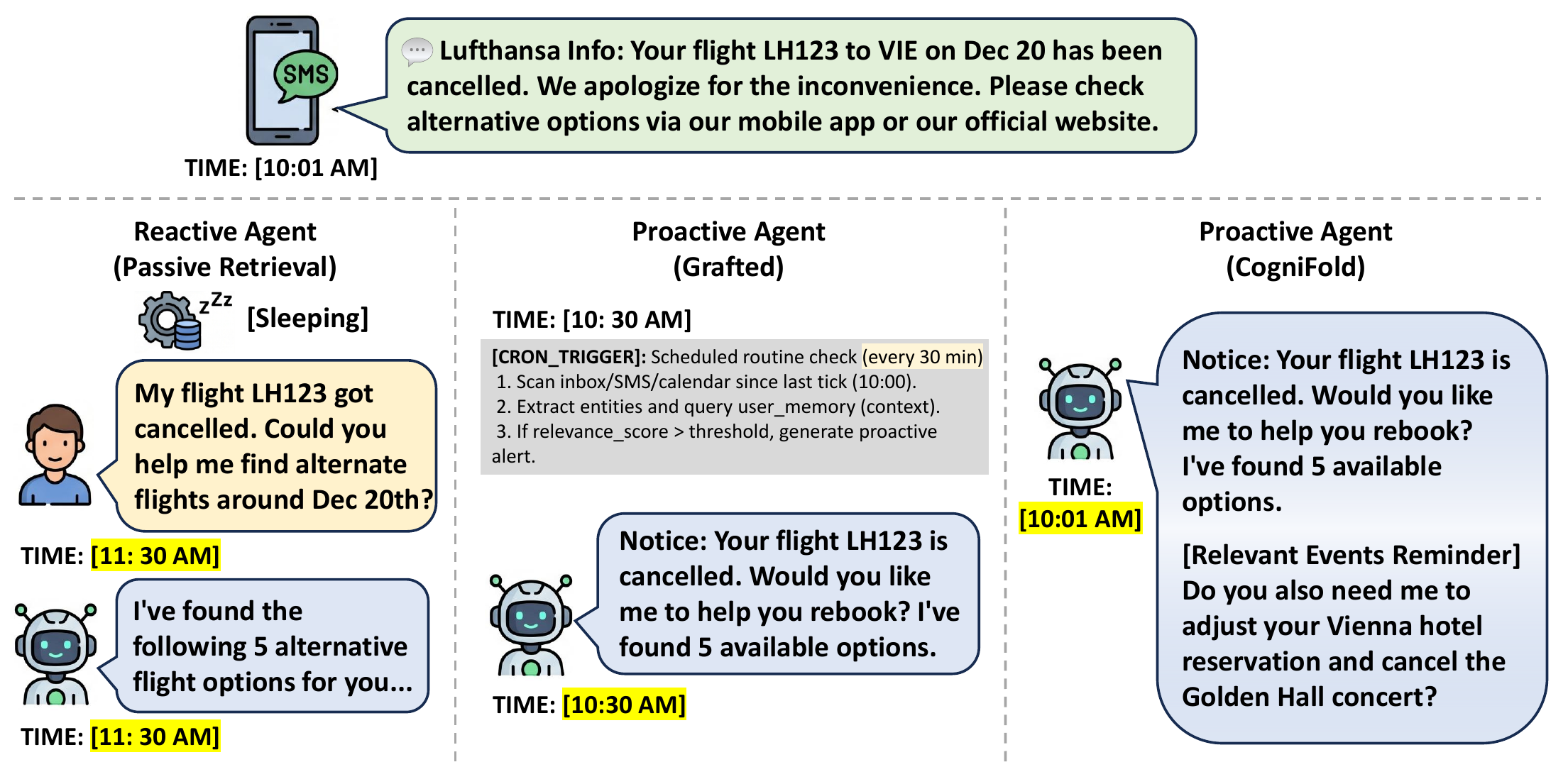}
\caption{\textbf{From reactive to proactive agent memory.} Conventional agents wait for explicit user queries (left) or graft delayed, application-layer triggers onto a reactive memory (middle). In contrast, \cognifold{} (right) processes unprompted, asynchronous events instantly within its memory substrate, simultaneously reactivating related dormant concepts (e.g., the Vienna hotel and concert).}
\label{fig:proactive_example}
\end{figure}

Human biological memory is evolutionarily adapted to exactly this setting: it continuously receives sensory input to autonomously encode, consolidate, forget, and surface intentions. 
Inspired by this, we propose \textbf{\cognifold{}, a proactive always-on agent memory} that folds continuously arriving events into self-emerging cognitive structure. \cognifold{} bootstraps in a strict sense: the graph's current state is the interpretive context for the next event, which in turn modifies the state for future events---a self-referential loop in which the system organises input entirely through its accumulated structure.  
\cognifold{} rests on two complementary perspectives.
From the \textbf{neural} side, we extend Complementary Learning Systems (CLS) theory~\citep{mcclelland1995there,kumaran2016learning} from two layers (hippocampus, neocortex) to three by adding a \textbf{Prefrontal Intent Layer}; rather than being hardcoded~\citep{bratman1987intention}, intents autonomously emerge once concept-cluster density crosses a threshold. From the \textbf{cognitive} side, the graph is a substrate for \textbf{conceptual bootstrapping}~\citep{carey2000origin, zhao2024model}: 
recursively scaffolding higher-level cognition from accumulated structure---a transparent, auditable form of test-time learning, distinct from both surface-level text rewriting (e.g., A-Mem~\citep{xu2025mem}) and opaque gradient updates (e.g., Titans~\citep{behrouz2024titans}).

We conduct a two-layer evaluation. At the \textbf{structural} layer, we introduce CogEval-Bench, a first-principles evaluation framework that directly measures whether the topology formed under continuous event streams matches cognitive expectations, demonstrating that \cognifold{} uniquely produces event-grounded concepts, coherent conceptual structure, and proactive intent emergence. At the \textbf{downstream} layer, we evaluate across eight benchmarks---two probing long-term conversational memory and six spanning other cognitive domains---confirming that \cognifold{} simultaneously performs competitively on conventional memory tasks.

Our contributions are summarized as follows:
\begin{itemize}[leftmargin=*, topsep=2pt, itemsep=2pt]
    \item \textbf{Always-On Proactive Memory Paradigm.} We recast agent memory from a reactive retrieval target into an always-on cognitive substrate (Fig.~\ref{fig:proactive_example}), natively supporting continuous understanding and proactive anticipation.
    \item \textbf{Tri-Layered Cognitive Architecture.} We extend the two-layer CLS framework with a prefrontal Intent layer, enabling self-emerging intents from accumulated concepts.
    \item \textbf{Continuous Topological Self-Organization.} We identify and algorithmically resolve four intrinsic structural debts of streaming events via transparent graph-level operations, yielding a transparent and auditable form of test-time learning.
    \item \textbf{CogEval-Bench Evaluation Framework.} We release a structural diagnostic evaluation framework that isolates proactive emergence from retrieval accuracy. Alongside eight established downstream benchmarks, we jointly validate \cognifold{}'s effectiveness in both high-level cognitive emergence and conventional memory robustness.
\end{itemize}

\section{CogniFold: From Neural Layers to Conceptual Bootstrapping}
\label{sec:architecture}

An always-on agent requires a fundamentally different memory substrate. Continuously arriving event streams demand an architecture capable of incremental, online integration. A genuinely autonomous assistant must transition from reactive retrieval to \textbf{proactive assembly}—continuously capturing implicit intents and organizing relevant cognitive structures in the background. \cognifold{} grounds this substrate in an extended Complementary Learning Systems (CLS) theory, formalizing memory as a dynamically evolving, typed multigraph.

\subsection{Tri-Layer Substrate}
\label{sec:neuro}

\begin{figure}[t]
    \centering
    \includegraphics[width=\textwidth]{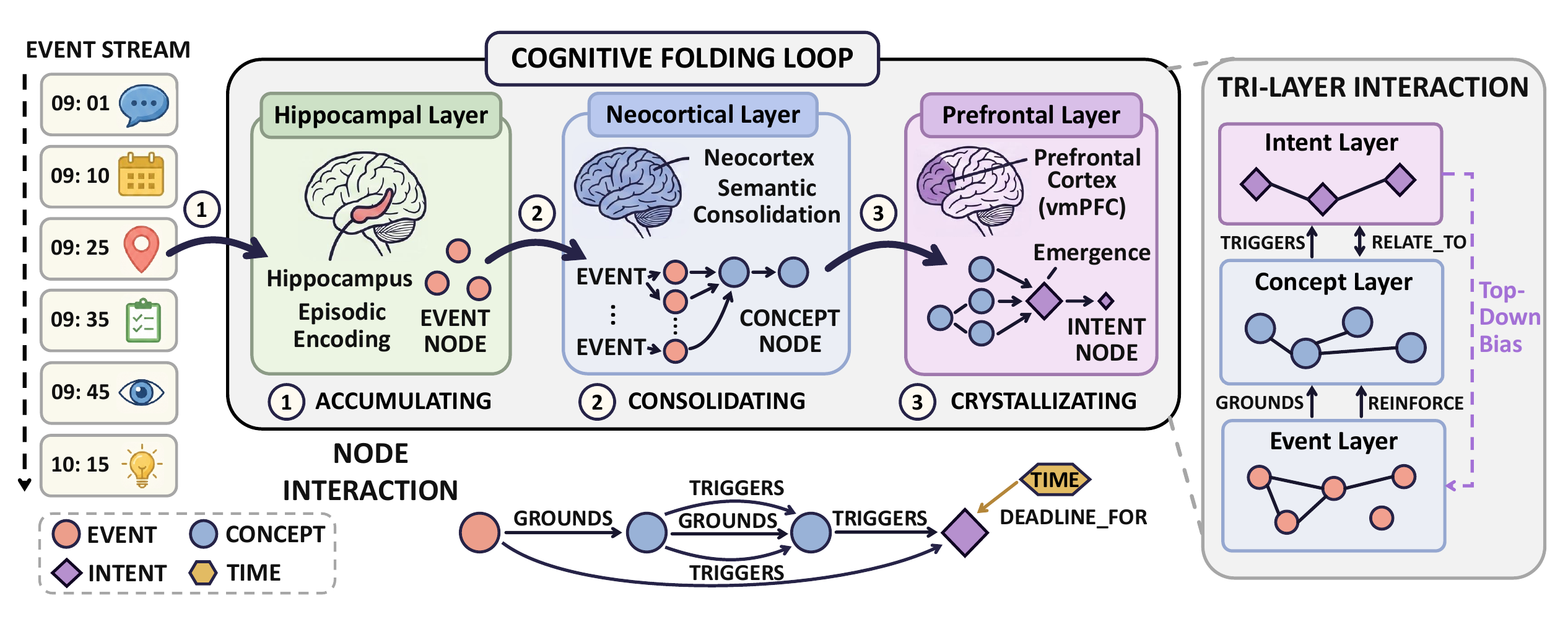}
    \caption{\textbf{The \cognifold{} Architecture: Conceptual Bootstrapping via Tri-Layered Cognitive Folding.} Extending the Complementary Learning Systems (CLS) framework, the memory substrate continuously metabolizes streaming events through three stages: accumulating raw episodic traces (Hippocampal layer), consolidating redundant patterns into semantic concepts (Neocortical layer), and crystallizing intents (Prefrontal layer). }
    \label{fig:example_image}
    
\end{figure}

Human declarative memory is organized by \textbf{Complementary Learning Systems} (CLS)~\citep{mcclelland1995there, kumaran2016learning}: the hippocampus rapidly encodes sparse, episode traces~\citep{marr1971simple, o1994hippocampal, squire1992memory}, while the neocortex slowly distills statistical regularities into semantic representations~\citep{tulving1972episodic, patterson2007you}.
This division is dynamic: over time, long-term storage shifts from the hippocampus to the medial prefrontal cortex (mPFC)~\citep{bontempi1999time, frankland2005organization}.

Crucially, the mPFC is not a passive recipient.
It exerts top-down control over the hippocampus via pre-existing knowledge frameworks (\textbf{schemata}) to actively shape which hippocampal traces are retained and how they are organized~\citep{tse2007schemas, de2026prefrontal}.
This bidirectional dialogue, in which the mPFC imposes schematic frameworks to guide subsequent encoding, forms the biological substrate from which goal-directed memory emerges~\citep{preston2013interplay, eichenbaum2017memory, van2012schema}.

\cognifold{} operationalizes the three-layer dialogue above as a typed, dynamically evolving multigraph.
\textbf{Event} nodes play the hippocampal role: each input from the stream is committed verbatim and time-stamped---an immutable episodic trace.
\textbf{Concept} nodes play the neocortical role: recurrent patterns are abstracted into schemata, anchored to their constituent events through provenance edges.
\textbf{Intent} nodes play the prefrontal role: when concept-level evidence converges into a coherent goal, an intent emerges and exerts top-down influence on how subsequent events are surfaced and encoded.

Yet, static layers are insufficient. Structure is merely the container of cognition; the vitality of memory lies in its \textbf{metabolism}. This brings us to the architectural dynamic at the core of \cognifold{}: \textbf{conceptual bootstrapping}.

\subsection{Dynamics: Conceptual Bootstrapping}
\label{sec:cog}

If the neuro perspective specifies the structural layers, conceptual bootstrapping~\citep{carey2000origin, zhao2024model} specifies how an agent ``pulls itself up by its own bootstraps'' on top of them. In \cognifold{}, this self-referential dynamic unfolds through \textbf{continuous folding} in three stages.

\textbf{Stage 1: Accumulation.}
The hippocampal layer (\textbf{Event} nodes) ingests the raw stream verbatim. Events initially function as cognitive \emph{placeholders}: raw experiential fragments committed before their overarching concepts exist.

\textbf{Stage 2: Consolidation.}
As events accumulate, the system detects statistical regularities across them and \emph{consolidates} them into the neocortical layer: discrete Event nodes are folded into \textbf{Concept} nodes anchored to their grounding events. 

\textbf{Stage 3: Crystallization.}
Concepts then act as active scaffolds for future input: incoming events are interpreted through them rather than from scratch. When concept-cluster density crosses a threshold, the bootstrap iterates upward—an \textbf{Intent} node \emph{crystallizes} in the prefrontal layer, providing top-down bias for schema-congruent encoding. The loop closes: structure interprets experience, and experience reshapes structure.

The synergy between neural structure and cognitive dynamic enables \cognifold{} to \textbf{metabolise} like biological memory: continuously folding to eliminate redundancy (compression) and bootstrapping to climb levels of abstraction, sustaining cognitive agility under an always-on event stream.

\subsection{Graph Formalization}
\label{sec:graph_def}

Having grounded \cognifold{} in neurobiological mapping (\S\ref{sec:neuro}) and cognitive dynamics (\S\ref{sec:cog}), we now formalise the substrate as a typed directed multigraph $\mathcal{G} = (\mathcal{V}, \mathcal{R})$ with four node types and nine semantic edge types (Table~\ref{tab:edge_types}).

\textbf{Node types.}
\textbf{Event} (episodic trace), \textbf{Concept} (semantic pattern), \textbf{Intent} (crystallized goal), and \textbf{Time} (temporal anchor).
The first three correspond to the CLS layers; \textbf{Time} is an auxiliary type connecting temporal obligations to intents via \texttt{DEADLINE\_FOR} edges.

\textbf{Edge ontology.}
Nine typed edges encode distinct semantic relations (Table~\ref{tab:edge_types}), each mapping to a specific cognitive motif.
This typed ontology constrains the LLM's update proposals toward meaningful topology, reducing the hallucination-driven bloat of free-form extraction.

\begin{table}[t]
\centering
\caption{\textbf{Edge types.} Each edge type maps to a specific cognitive/biological motif (CLS Analogue column). Default weights are reported in Appendix~\ref{app:hyperparams}.}
\label{tab:edge_types}
\small
\begin{tabular}{@{}lll@{}}
\toprule
Edge Type & Semantics & CLS Analogue \\
\midrule
\textsc{Grounds} & Event evidences concept/intent & Episodic grounding~\citep{mcclelland1995there} \\
\textsc{Causes} & Event causes event & Causal encoding \\
\textsc{Triggers} & Concept triggers intent & Goal activation~\citep{einstein2005prospective} \\
\textsc{Reinforces} & Event supports concept & Synaptic LTP~\citep{hebb2005organization} \\
\textsc{Part\_Of} & Structural hierarchy & Hierarchical binding~\citep{eichenbaum2017memory} \\
\textsc{Derived\_From} & Concept abstraction & Schema generalization~\citep{bartlett1995remembering} \\
\textsc{Deadline\_For} & Temporal constraint & Prospective memory~\citep{einstein2005prospective} \\
\textsc{Related\_To} & Associative link & Associative memory~\citep{mcclelland1995there} \\
\textsc{User\_Feedback} & Feedback $\to$ intent & Valence signal \\
\bottomrule
\end{tabular}
\end{table}

\textbf{Write/read decoupling.}
The architecture decouples graph expansion from query execution.
The write path (\S\ref{sec:write_path}, \S\ref{sec:intent}) specifies topology-evolution operations that run on every incoming event; the read path (\S\ref{app:query}) specifies multi-strategy retrieval over the graph snapshot (parameters in Appendix~\ref{app:hyperparams}). This ensures formation and retrieval can be diagnosed independently.

\section{Continuous Cognitive Folding}
\label{sec:mechanism}

\begin{figure}[t]
    \centering
    \includegraphics[width=\textwidth]{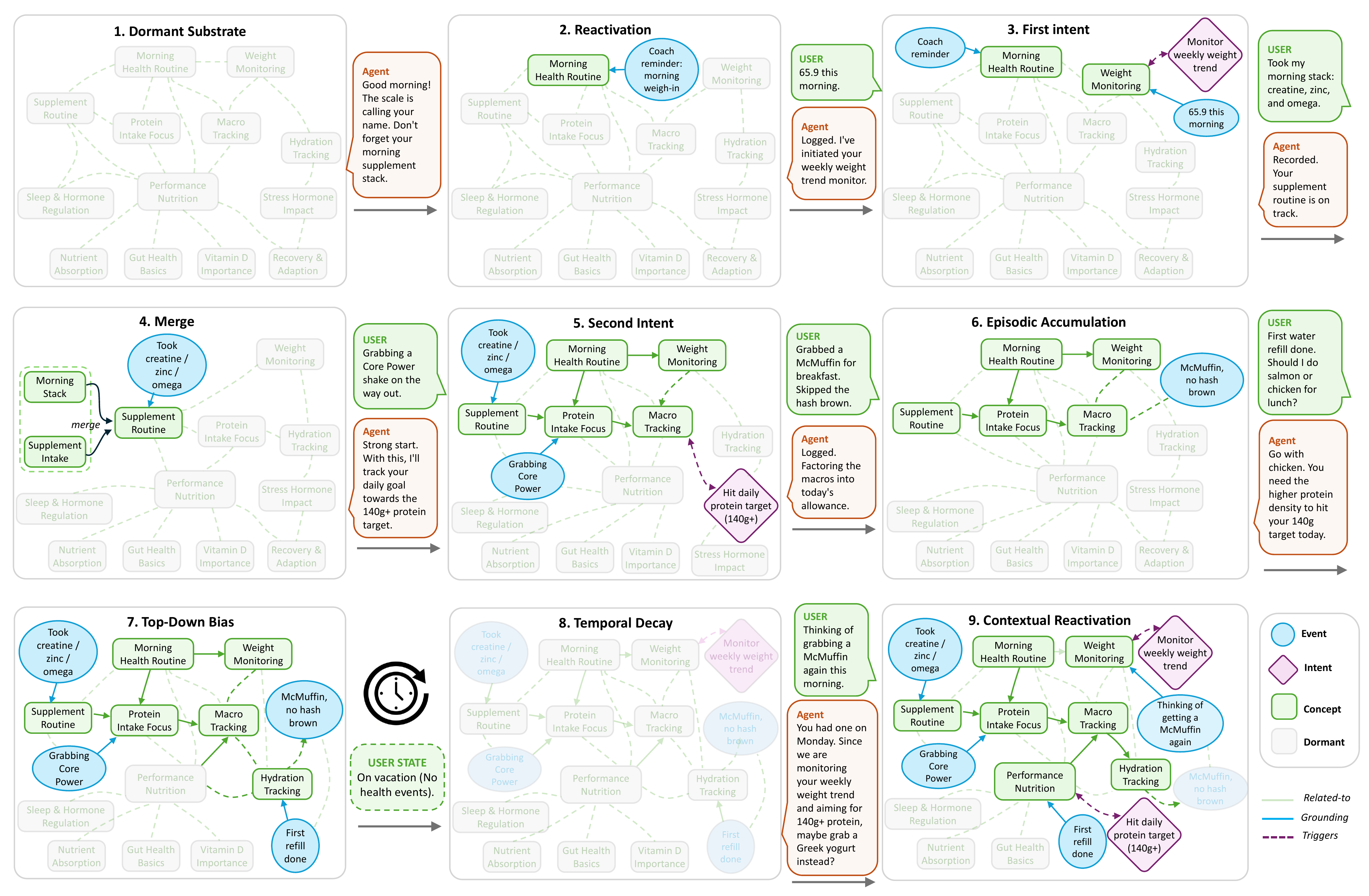}
    \caption{\textbf{Continuous cognitive metabolism.} Under an asynchronous event stream, the memory substrate dynamically self-organizes. The graph autonomously consolidates episodic events (Panel 3), merges associated schemata (Panel 4), and crystallizes goal-directed intents from converging concept density (Panel 5). This living topology natively supports top-down cognitive bias (Panel 7), natural temporal decay (Panel 8), and structure-driven proactive intervention (Panel 9).}
\label{fig:metabolism_cycle}
\end{figure}

Reactive memory architectures enjoy considerable design slack: ingestion is bound to user turns, consolidation can be deferred offline, and retrieval is the only operation under latency pressure.
A proactive, always-on agent has none of these.
Events arrive continuously and asynchronously, working memory stays bounded, and the next query may concern structure that has not yet been formed---all between user touchpoints.
The graph must therefore mutate in place under a stream that never pauses; topology must keep paying down the four structural debts---\textbf{accumulation, compression, decay, completion}---that any continuously evolving graph naturally accrues; and a proactive agent must assemble relevant context before being asked, which forces structural centrality, temporal recency, and usage intensity to be treated as simultaneous hard constraints rather than retrieval-time heuristics.

Three mechanisms purpose-built under these pressures \textbf{operationalise conceptual bootstrapping (\S\ref{sec:cog}) under the stream}: a proactive context-assembly harness on the write path (\S\ref{sec:write_path}), automatic topology-evolution operations that discharge the four debts (\S\ref{sec:debts}), and an intent-emergence stage that crystallizes goals from converging concept evidence (\S\ref{sec:intent}).

\subsection{Proactive Context Assembly}
\label{sec:write_path}

The working-memory constraint is Miller's classical capacity bound~\citep{miller1956magical} transposed to an agent: each encoding step can only reason over a tiny subset of accumulated knowledge. This imposes a \textbf{priority allocation problem} on the write path: the system must select which subset of the existing graph the LLM sees when interpreting the next event.

Priority is allocated through three signals, each anchored in a distinct cognitive memory tradition.

\textbf{Structural centrality}---how embedded a node is in the cognitive graph---follows the Personalized PageRank tradition~\citep{gutierrez2024hipporag}; we extend it from retrieval into the write path because what the LLM sees during encoding directly shapes what it writes, making centrality a formation-time prior rather than only a query-time signal.

\textbf{Temporal recency}---whether a trace is still fresh---follows the Ebbinghaus forgetting curve~\citep{ebbinghaus2013image}, applied to LLM memory in MemoryBank~\citep{zhong2024memorybank}; the same exponential kernel governs our write-path priority so that stale schemata do not perpetually crowd out new evidence.

\textbf{Access intensity}---how often a node has been re-engaged---is a Hebbian signal~\citep{hebb2005organization}: nodes that repeatedly co-fire with the agent's working context wire more strongly into the next context, akin to the access-count heuristic in Mem0~\citep{chhikara2025mem0} but lifted from a passive tally to an active scoring term.

These three signals compose linearly into a per-node priority:
\begin{equation}
    \text{Score}(v) = \bigl[\alpha \cdot \text{PR}(v) + \beta \cdot \exp(-\lambda \cdot \Delta t_v) + \gamma \cdot \text{Acc}(v)\bigr] \cdot U(v)
    \label{eq:score}
\end{equation}
where $U(v) \geq 1$ is a deadline-driven urgency multiplier from connected \textbf{Time} nodes; weights are reported in Appendix~\ref{app:hyperparams}.

The resulting scores define a \textbf{proactive context window}: structurally central, temporally fresh, and frequently used knowledge surfaces before the next event is interpreted---rather than waiting for a later query to reveal what should have mattered. Selected nodes are partitioned into Immediate, Working, and Background tiers (proportions in Appendix~\ref{app:hyperparams}), forcing the LLM to process $e_t$ within a layered subgraph rather than in isolation.

Given the assembled context, an LLM central executive emits an \texttt{UpdatePlan}: a sequence of atomic operations (\texttt{ADD\_NODE}, \texttt{ADD\_EDGE}, \texttt{UPDATE\_NODE}, \texttt{MERGE\_NODES}, \texttt{REMOVE\_NODE}), each carrying natural-language reasoning and \texttt{grounded\_in} provenance.
The executor validates and applies the plan atomically with snapshot-based rollback, and near-duplicate concepts (above a title-similarity threshold; see Appendix~\ref{app:hyperparams}) are silently converted to reinforcement updates to prevent bloat from redundant events.

\subsection{Four Structural Debts}
\label{sec:debts}

Continuous event arrival is not a neutral inflow: by the nature of the input, a memory graph accumulates four kinds of structural debt over time.
These are not design choices but mandatory state-change operations imposed by the stream; any always-on memory that fails to address any one of them degrades along the corresponding axis.
\cognifold{} addresses all four as automatic graph-level operations, executed without per-step LLM supervision in a consolidation pass inspired by sleep-dependent consolidation~\citep{stickgold2007sleep}.

\begin{enumerate}[leftmargin=*, topsep=2pt, itemsep=2pt]
\item \textbf{Accumulation}---persistent patterns must strengthen; one-off noise must not.
Without it, a one-off event and a recurring concept reach equivalent PageRank.
Our operation: when a new event corroborates an existing concept, the system creates a \texttt{REINFORCES} edge rather than a duplicate node, boosting that concept's in-degree and PageRank---implementing Bartlett's schema assimilation~\citep{bartlett1995remembering} as a graph operation.

\item \textbf{Compression}---redundant fragments must fold.
Without it, graph size grows with $|\text{events}|$, PageRank diffuses across duplicates, and evidence that should aggregate stays fragmented.
Our operation: when two concept nodes exceed a semantic-similarity threshold (Appendix~\ref{app:hyperparams}), the executor automatically merges them (\texttt{MERGE\_NODES}); the higher-access node absorbs all edges.
This implements schema unitization~\citep{gilboa2017neurobiology} and physically shortens graph-theoretic distances---multi-hop chains collapse to direct adjacency.

\item \textbf{Decay}---aged structure must weaken.
Without it, there is no forgetting; stale connections dominate attention, and ``recent'' becomes indistinguishable from ``active''.
Our operation: all edges undergo exponential decay at every consolidation pass, following MemoryBank's~\citep{zhong2024memorybank} application of the Ebbinghaus curve.

\item \textbf{Completion}---connections invisible to a local LLM view must be inferred.
Without it, the LLM sees only the current event plus its context window and cannot know that a concept created now should connect to one created three sessions ago; cross-session structure fragments into orphans.
Our operation: kNN inference over concept embeddings (parameters in Appendix~\ref{app:hyperparams}) scans for zero-edge concept nodes and creates \texttt{GROUNDS} connections---automatically repairing gaps the LLM's local-view planning misses.
\end{enumerate}

Prior systems address each debt at most partially (Appendix~\ref{app:attack_surface}, Table~\ref{tab:attack_surface}): HippoRAG covers a narrow form of completion via synonym edges; Mem0 performs write-time dedup without post-hoc consolidation or decay; MAGMA's slow-path inference densifies an ingested batch but does not reinforce, compress, or decay; A-Mem and PREMem operate at the text-rewrite layer and never modify graph structure.
\cognifold{} is the first agent-memory system to address all four debts as automatic, topology-level operations---the mutually-reinforcing cycle \texttt{REINFORCES} $\to$ \texttt{MERGE\_NODES} $\to$ kNN completion, balanced by edge decay.
Why graph-level and not text-level or gradient-level?
Text rewriting updates the content of a note while leaving graph-theoretic distance, PageRank, and reasoning paths invariant; gradient-based memory updates weights that cannot be inspected, audited, or selectively deleted~\citep{mccloskey1989catastrophic}.
Only topology change makes memory's internal geometry both mutable and inspectable.

\subsection{Intent Emergence}
\label{sec:intent}

\textbf{Intent} nodes emerge when concept-cluster density crosses a threshold~\citep{einstein2005prospective, gilboa2017neurobiology}: converging evidence across multiple concepts signals an unmet goal, and the LLM crystallizes it as an intent linked to its supporting concepts via \texttt{TRIGGERS} edges.
Each intent follows a lifecycle (\texttt{pending} $\to$ \texttt{resolved} $|$ \texttt{rejected} $|$ \texttt{deferred}) that provides goal-directed organization for always-on agents.

A per-category EMA loop calibrates the emission threshold from accept/reject/defer/modify feedback:
\begin{equation}
    w_c^{(t)} = (1 - \alpha_{\text{ema}}) \cdot w_c^{(t-1)} + \alpha_{\text{ema}} \cdot s_t,
    \label{eq:ema}
\end{equation}
where $s_t$ maps each feedback type to a numeric score (Appendix~\ref{app:hyperparams}).
Categories the user consistently accepts see lowered thresholds; rejected categories are suppressed---prediction-error correction~\citep{friston2010free, clark2013whatever} applied to intent generation rather than model weights.

Under the downstream QA protocols of \S\ref{sec:experiments}, concept-cluster density never reaches the emission threshold, so intents are not triggered there.
In the controlled multi-domain streams of CogEval-Bench (\S\ref{sec:cogeval_rq}), the threshold is reached repeatedly and intent emission is measured directly (Proactivity~0.614).

\section{Experiments and Results}
\label{sec:experiments}

\begin{table}[t]
\centering
\caption{\textbf{Downstream benchmark suite.} Eight benchmarks: two on long-term conversational memory and six across other cognitive domains. Detailed system comparisons are in Table~\ref{tab:locomo_main} (LoCoMo), Table~\ref{tab:longmemeval_main} (LongMemEval), and Figure~\ref{fig:benchmark_bars} (the other six).}
\label{tab:downstream_overview}
\small
\begin{tabular}{@{}lllc@{}}
\toprule
Benchmark & Domain & Metric & $n$ \\
\midrule
LoCoMo~\citep{maharana2024evaluating}     & Conversational memory       & J-Score / F1 & 1540 \\
LongMemEval~\citep{wu2025longmemeval} & Conversational memory       & LLM-judge acc.\                    & 500  \\
MuSiQue~\citep{trivedi2022musique}    & Multi-hop reasoning         & EM / F1                            & 500  \\
NarrativeQA~\citep{kovcisky2018narrativeqa} & Narrative comprehension & F1                                 & 500  \\
StreamingQA~\citep{liska2022streamingqa}    & Streaming temporal QA   & F1                                 & 500  \\
MuTual~\citep{cui2020mutual}        & Dialogue coherence          & Accuracy                           & 500  \\
ToMi~\citep{le-etal-2019-revisiting}     & Theory of mind              & Exact match                        & 500  \\
BABILong~\citep{kuratov2024babilong}  & Long-context factual extr.\ & Exact match                        & 100  \\
\bottomrule
\end{tabular}
\end{table}

\subsection{Datasets}
\label{sec:exp_datasets}

\textbf{Proactive Evaluation.}
QA accuracy alone cannot validate the central claim that cognitive structures emerge from event-stream folding---a flat RAG system with strong BM25 can score well on factual QA without forming any concepts, and a verbatim event store can pass multi-hop retrieval without any compression.
We therefore introduce \textbf{CogEval-Bench}, a structural diagnostic benchmark.\footnote{Dataset available at \url{https://huggingface.co/datasets/OpenUploading/CogEval-Bench}.}
CogEval-Bench uses \textbf{top-down generation}: for each scenario a gold concept graph $\mathcal{G}^* = (\mathcal{C}^*, \mathcal{R}^*, \mathcal{H}^*, \mathcal{I}^*)$ is specified first (concepts, inter-concept relations, hierarchy parents, expected intents, and planted multi-hop chains), then grounded first-person events are generated from it, followed by distractor injection (10--15\%) and temporal shuffling.
The benchmark spans 6 scenarios across 4 domains (SoftEng, Health, Team, News, Academic, Support); scale statistics are reported in Appendix~\ref{app:cogeval_details}.
Ground truth is established \textbf{by construction} rather than through post-hoc annotation.
Three evaluation tracks are computed per system: \textbf{Concept Emergence} (Gold~F1 via Hungarian-matched~\citep{kuhn1955hungarian} soft-matching, LLM Quality, Harmony, Purity), \textbf{Relationship Topology} (Chain Discovery, Clustering, Modularity, Edge Type Entropy), and \textbf{Compression \& Proactivity} (Compression Ratio, PageRank Gini, Proactivity).
Full schemas, generation prompts, and per-scenario breakdowns are in Appendix~\ref{app:cogeval_details}.

\textbf{Memory-Quality Evaluation.}
We evaluate downstream memory utility on two conversational-memory benchmarks and six broader probes. The conversational pair is \textbf{LoCoMo}~\citep{maharana2024evaluating} (full 10-conversation Mem0 protocol) and \textbf{LongMemEval}~\citep{wu2025longmemeval} ($500$ questions over ${\sim}50$-session histories, which adds the knowledge-update and abstention cases LoCoMo leaves out). The other six cover separate cognitive domains: dialogue coherence (\textbf{MuTual}~\citep{cui2020mutual}), theory of mind (\textbf{ToMi}~\citep{le-etal-2019-revisiting}), multi-hop reasoning (\textbf{MuSiQue}~\citep{trivedi2022musique}), narrative comprehension (\textbf{NarrativeQA}~\citep{kovcisky2018narrativeqa}), streaming temporal QA (\textbf{StreamingQA}~\citep{liska2022streamingqa}), and long-context factual extraction (\textbf{BABILong}~\citep{kuratov2024babilong}). Per-benchmark sample sizes, baselines, and detailed results are in \S\ref{sec:exp_downstream} (Table~\ref{tab:downstream_overview} and Figure~\ref{fig:benchmark_bars}).

\subsection{Baselines}
\label{sec:exp_baselines}

\textbf{Memory-Quality baselines.}
On LoCoMo, we compare against MIRIX~\citep{wang2025mirix}, Mem0~\citep{chhikara2025mem0}, Zep~\citep{rasmussen2025zep}, Memobase~\citep{memobase2024}, Supermemory~\citep{supermemory2024}, MemU~\citep{memu2025}, MemOS~\citep{li2025memos}, and ENGRAM~\citep{patel2025engram} under the matched single-judge \texttt{gpt-4o-mini} Mem0 protocol; numbers come from~\citet{li2025memos}'s public reproduction for all but ENGRAM (taken from~\citealp{patel2025engram}) and Zep (corrected reproduction~\citep{rasmussen2025zep}).
On MuSiQue, we adopt the standard graph-retrieval suite of~\citet{gutierrez2025rag}---BM25, Contriever, NV-Embed-v2, RAPTOR~\citep{sarthi2024raptor}, GraphRAG~\citep{edge2024local}, LightRAG~\citep{guo2024lightrag}, HippoRAG~\citep{gutierrez2024hipporag}, HippoRAG\,2~\citep{gutierrez2025rag}---plus PolicyRAG~\citep{sarnaikpolicyrag}.
On the remaining benchmarks we report against the most-cited published baselines under each benchmark's headline metric (Figure~\ref{fig:benchmark_bars}).

\textbf{Proactive baselines.}
On CogEval-Bench, seven systems are compared under identical LLM and events: \textbf{OpenIE KG} (HippoRAG-style triples), \textbf{Cognee}~\citep{cognee2024} (Extract--Cognify--Load pipeline), \textbf{HippoRAG\,2}~\citep{gutierrez2025rag} (deep passage + synonym expansion), \textbf{GraphRAG}~\citep{edge2024local} (batch community detection + LLM summarisation), \textbf{Mem0}~\citep{chhikara2025mem0} (text-rewrite memory cells), \textbf{Zep}~\citep{rasmussen2025zep} (temporal knowledge graph), and \cognifold{}.

\subsection{Implementation Details}
\label{sec:exp_impl}
\begin{table}[t]
\centering
\caption{\textbf{CogEval-Bench: structural evaluation across 7 systems.} Averages over 6 scenarios (small scale, ${\sim}42$ events each). Track~A measures concept quality, Track~B measures graph topology, Track~C measures compression and proactivity. Arrows indicate preferred direction. All systems share GPT-4o-mini and \texttt{text-embedding-3-small}; differences are attributable to architecture. Only \cognifold{} achieves non-zero purity and proactivity---structural properties absent from entity-level, batch-processed, or text-rewrite representations. Bold: best; underline: second-best.}
\label{tab:cogeval_main}
\small
\setlength{\tabcolsep}{4pt}
\begin{tabular}{@{}llccccccc@{}}
\toprule
Track & Metric & OpenIE KG & Cognee & HippoRAG\,2 & GraphRAG & Mem0 & Zep & \cognifold{} \\
\midrule
\multirow{4}{*}{\rotatebox[origin=c]{90}{\scriptsize Track A}}
& Harmony $\uparrow$ & 0.138 & 0.094 & 0.095 & \underline{0.323} & 0.000 & 0.138 & \textbf{0.476} \\
& Gold F1 $\uparrow$ & 0.081 & 0.053 & 0.058 & \underline{0.232} & 0.000 & 0.081 & \textbf{0.358} \\
& LLM Quality $\uparrow$ & 0.492 & 0.454 & 0.341 & \underline{0.541} & 0.000 & 0.494 & \textbf{0.733} \\
& Purity $\uparrow$ & 0.000 & 0.000 & 0.000 & 0.000 & 0.000 & 0.000 & \textbf{0.361} \\
\midrule
\multirow{4}{*}{\rotatebox[origin=c]{90}{\scriptsize Track B}}
& Chain Disc. $\uparrow$ & 0.750 & \textbf{1.000} & \underline{0.917} & 0.833 & \textbf{1.000}$^\dagger$ & \underline{0.917} & 0.833 \\
& Clustering $\uparrow$ & 0.005 & 0.200 & \textbf{0.716} & 0.002 & 0.382 & \underline{0.704} & 0.327 \\
& Modularity & 0.848 & 0.714 & 0.759 & \textbf{0.873} & 0.434 & 0.784 & 0.546 \\
& Edge Entropy $\uparrow$ & \textbf{0.968} & 0.540 & 0.000 & \underline{0.968} & 0.000 & 0.507 & 0.624 \\
\midrule
\multirow{3}{*}{\rotatebox[origin=c]{90}{\scriptsize Track C}}
& Compression $\uparrow$ & 0.3$\times$ & 0.5$\times$ & 0.2$\times$ & \underline{1.2$\times$} & 1.0$\times$ & 0.5$\times$ & \textbf{4.6$\times$} \\
& PR Gini & 0.323 & 0.352 & 0.310 & 0.317 & 0.189 & 0.242 & 0.274 \\
& Proactivity $\uparrow$ & 0.000 & 0.000 & 0.000 & 0.000 & 0.000 & 0.000 & \textbf{0.614} \\
\bottomrule
\end{tabular}

\smallskip
{\scriptsize $^\dagger$Mem0 has no native graph; we materialise each extracted memory as a node and induce edges via vector-similarity neighbours, which trivially saturates Chain~Disc.\ but produces no typed-edge structure (Edge~Entropy~$=0$).}
\end{table}

The agent is \texttt{gpt-4o-mini} on every benchmark, and so is the reader except on LongMemEval, which answers with \texttt{gpt-5.4-mini} (\S\ref{sec:exp_downstream}); embeddings are \texttt{text-embedding-3-small} throughout. Since every system builds its memory with the same small agent, what separates them is the representation, not the extraction model. All benchmarks run with stream ingestion: each event is processed sequentially through the full write path (context assembly $\to$ \texttt{UpdatePlan} $\to$ atomic execution $\to$ consolidation), so the topology-evolution operations of \S\ref{sec:debts} fire per event and consolidation operates throughout ingestion---the same online operation \cognifold{} performs in deployment, not a one-shot batch pass at the end. Hyper-parameters, full prompts, and the cost / reproducibility statement are in Appendix~\ref{app:hyperparams} and Appendix~\ref{app:reproducibility}.

\subsection{Results}

\textbf{Proactive Results.}
\label{sec:cogeval_rq}
Table~\ref{tab:cogeval_main} reports CogEval-Bench averages across all six scenarios; per-scenario breakdowns and full evaluation details are in Appendix~\ref{app:cogeval_details}.
\cognifold{} achieves Harmony 0.476, substantially above GraphRAG (0.323, the strongest baseline) and far above entity-level systems (OpenIE KG 0.138; Cognee 0.094; HippoRAG\,2 0.095).
\cognifold{} is the only system producing non-zero Purity (0.361)---its concepts are coherently grounded in their constituent events, while all baselines lack event-level grounding.
On topology, \cognifold{}'s clustering coefficient (0.327) reflects genuine triadic closure among semantically related concepts, distinct from HippoRAG\,2's high raw clustering (0.716) that arises from synonym-expansion cliques among name variants.
On compression and proactivity, \cognifold{} achieves $4.6\times$ compression (41--47 events $\to$ 7--12 concepts) while OpenIE KG and HippoRAG\,2 expand the representation; and \cognifold{} is the only system emitting intent nodes at all, reaching Proactivity~0.614 (61\% of intents grounded by $\geq 2$ supporting connections).

The seven-system comparison reveals an ordered hierarchy in representational richness:
\textbf{entity graphs} (OpenIE KG)---fragmented triples, high modularity from disconnection;
\textbf{entity graphs with enrichment} (Cognee, HippoRAG\,2)---shallow structure atop entity extraction, Harmony stuck at ${\sim}0.09$;
\textbf{community graphs} (GraphRAG)---batch community detection yields the strongest baseline (Harmony 0.323) but with zero event-level grounding (Purity~$=0$) and negligible clustering (0.002);
\textbf{cognitive graphs} (\cognifold{})---online folding with merging yields event-grounded concepts, genuine triadic closure, substantial compression, and proactive goal identification simultaneously.
The critical architectural distinction is \textbf{online, incremental processing with merging}: Cognee and HippoRAG\,2 add machinery (ECL, synonym expansion, PPR) without cross-event integration and remain at the entity-enrichment tier, confirming that the bottleneck in concept emergence is not extraction depth but the ability to recognise that events $e_1, e_3, e_7$ ground the same underlying concept and merge them into a single abstraction.
This hierarchy parallels a neuroscience progression~\citep{mcclelland1995there, gilboa2017neurobiology, preston2013interplay}: episodic storage, pattern separation without consolidation, shallow categorisation, schema extraction, and active consolidation with goal generation.

\textbf{Memory-Quality Results.}
\label{sec:exp_downstream}
CogEval-Bench showed the substrate forms cognitive structure; the eight downstream benchmarks (Table~\ref{tab:downstream_overview}) test whether it pays off as memory. Two are conversational---LoCoMo and LongMemEval---and we put them against dedicated memory systems. The other six reach into separate cognitive domains, each judged against that domain's own published baselines. The graph never changes between them: the write path builds it once, the read path queries it.

\textit{Long-term conversational memory.}
\begin{table}[t]
\centering
\caption{\textbf{LoCoMo per-category and aggregate comparison.} J-Score is the LLM-as-judge accuracy~\citep{zheng2023judging} with \texttt{gpt-4o-mini} as the judge; Tokens is per-question input+output token consumption. \textbf{Bold}: best per column; \underline{underline}: second-best.}
\label{tab:locomo_main}
\small
\setlength{\tabcolsep}{4pt}
\begin{tabular}{@{}lccccccc@{}}
\toprule
Model & Tokens & Single-Hop & Multi-Hop & Temporal & Open Domain & \textbf{Overall} & Overall F1 \\
\midrule
MIRIX~(\citeyear{wang2025mirix})         & --   & 68.22 & 54.26 & 68.54 & 46.88 & 64.33 & 28.10 \\
Mem0~(\citeyear{chhikara2025mem0})      & 1.2k & 67.13 & 51.15 & 55.51 & \textbf{72.93} & 66.88 & 43.46 \\
Zep~(\citeyear{rasmussen2025zep})  & 2.7k & 79.79 & 74.11 & 67.71 & 66.04 & 75.14 & 41.23 \\
Memobase~(\citeyear{memobase2024})  & 2.1k & 73.12 & 64.65 & \underline{81.20} & 53.12 & 72.01 & \textbf{50.18} \\
Supermemory~(\citeyear{supermemory2024}) & 0.5k & 67.30 & 51.12 & 31.77 & 42.67 & 55.34 & 34.87 \\
MemU~(\citeyear{memu2025})               & 0.6k & 66.34 & 63.12 & 27.10 & 50.00 & 56.55 & 35.15 \\
MemOS~(\citeyear{li2025memos})           & 2.6k & 81.09 & 67.49 & 75.18 & 55.90 & 75.80 & \underline{45.27} \\
ENGRAM~(\citeyear{patel2025engram})           & 0.9k & 79.90 & \underline{79.79} & 70.79 & \underline{72.92} & 77.55 & 21.08 \\
EverMemOS$^\dagger$~(\citeyear{hu2026evermemos}) & 2.5k & \textbf{91.08} & \textbf{86.17} & \textbf{81.93} & 66.67 & \textbf{86.76} & --    \\
\midrule
\textbf{\cognifold{}}            & 4.2k & \underline{90.49} & 67.38 & 78.50 & 50.00 & \underline{81.23} & 35.71 \\
\bottomrule
\end{tabular}
\\[4pt]
{\scriptsize $^\dagger$Reported under a 3-LLM-judge ensemble protocol; numbers as reported in~\citet{hu2026evermemos} Table 1.}
\end{table}

On LoCoMo (Table~\ref{tab:locomo_main}, full 10-conversation Mem0 protocol with matched judge), the substrate is set against eight memory systems and leads the audit-resilient region of the leaderboard~\citep{penfield2026locomoaudit}, scoring above MemOS, ENGRAM, and the text-rewriting tier.
\begin{table}[t]
\centering
\caption{\textbf{LongMemEval-S comparison.} LLM-as-judge accuracy (\%)~\citep{zheng2023judging} on the $500$-question split of~\citet{wu2025longmemeval}. \emph{Build} and \emph{Answer} are the models on the write (memory construction) and read (QA) paths. \textbf{Bold}/\underline{underline}: best/second per column.}
\label{tab:longmemeval_main}
\setlength{\tabcolsep}{4pt}
\resizebox{\textwidth}{!}{%
\begin{tabular}{@{}llllccccccc@{}}
\toprule
Model & Build & Answer & Judge & SSU & SSA & SSP & MS & KU & TR & \textbf{Overall} \\
\midrule
Mem0$^\dagger$~(\citeyear{chhikara2025mem0})       & 4o-mini    & 4o-mini  & 4o-mini  & 81.4 & 41.1 & 60.0 & 46.2 & 70.1 & 40.2 & 53.6 \\
LightMem~(\citeyear{fang2025lightmem})             & 4o-mini    & 4o-mini  & 4o-mini  & 87.1 & 32.1 & 68.2 & 71.7 & 83.1 & 67.2 & 68.6 \\
Zep~(\citeyear{rasmussen2025zep})                  & 4o-mini    & 4o       & 4o       & 92.9 & 80.4 & 56.7 & 57.9 & 83.3 & 62.4 & 71.2 \\
ENGRAM~(\citeyear{patel2025engram})                & 4o-mini    & 4o-mini  & 4o-mini  & \underline{97.1} & 87.5 & \underline{93.3} & 60.2 & 74.4 & 55.6 & 71.4 \\
SimpleMem~(\citeyear{liu2026simplemem})            & 4.1-mini   & 4.1-mini & 4.1-mini & 85.7 & 75.0 & 76.7 & 60.9 & 79.5 & 83.5 & 76.9 \\
EverMemOS$^\ddagger$~(\citeyear{hu2026evermemos})  & 4.1-mini   & 4.1-mini & 3-ens.   & \underline{97.1} & 85.7 & \underline{93.3} & 73.7 & 89.7 & 77.4 & 83.0 \\
Supermemory$^\S$~(\citeyear{supermemory2024})      & 4o         & 5        & 4o       & \underline{97.1} & \textbf{100.0} & 76.7 & 75.2 & 87.2 & 81.2 & 84.6 \\
Chronos (Low)$^\P$~(\citeyear{chronos2026})        & --         & 4o       & 4o       & 94.3 & \textbf{100.0} & 80.0 & \textbf{91.7} & \underline{96.2} & \underline{90.2} & 92.6 \\
Chronos (High)$^\P$                                & --         & Opus-4.6 & 4o       & \textbf{98.6} & \textbf{100.0} & \textbf{100.0} & 88.7 & \textbf{100.0} & \textbf{95.5} & \textbf{95.6} \\
Mastra$^\S$~(\citeyear{mastra2026om})              & gemini-2.5-flash & 5-mini & 4o & 95.7 & \underline{94.6} & \textbf{100.0} & 87.2 & \underline{96.2} & \textbf{95.5} & \underline{94.9} \\
\midrule
\textbf{\cognifold{}}                              & 4o-mini & 5.4-mini & 4o & \underline{97.1} & \textbf{100.0} & \underline{93.3} & \underline{91.0} & 94.9 & 88.7 & 93.0 \\
\bottomrule
\end{tabular}%
}
\\[3pt]
{\scriptsize $^\dagger$Mem0 has no first-party LongMemEval result; row from~\citet{fang2025lightmem}. $^\ddagger$3-LLM-judge ensemble. $^\S$Vendor-reported, not peer-reviewed. $^\P$\textsc{Chronos}~\citep{chronos2026} does not name its extraction model; \emph{Answer} is its generation model only, and it retrieves with \texttt{text-embedding-3-large}. Names drop \texttt{gpt-}/\texttt{gemini-} prefixes; \texttt{Opus-4.6} is Claude~Opus~4.6.}
\end{table}

LongMemEval~\citep{wu2025longmemeval} stresses the same conversational memory far harder, and it is where the gap between a cheap write path and a strong read path shows most. Its $500$ questions cover single-session recall, multi-session consolidation, knowledge update, and temporal reasoning over histories of ${\sim}50$ sessions (Table~\ref{tab:longmemeval_main}); here the real cost is ingesting every turn around the clock, not producing the one-shot answer. \cognifold{} reaches $93.0\%$ overall and ties the field's best on single-session-assistant recall (SSA~$100.0$), third behind Chronos's Claude-Opus-4.6 configuration ($95.6$) and Mastra ($94.9$). But the ranking is the wrong axis to read the table on; the controlled comparison is the \emph{Build} column. Five systems hold memory construction to \texttt{gpt-4o-mini}---Mem0, LightMem, Zep, ENGRAM, and \cognifold{}---and among them \cognifold{} leads by more than twenty points ($93.0$ vs.\ the next-best $71.4$). Where the top configuration pulls ahead, it pays for it at exactly the stages \cognifold{} keeps cheap: Chronos~High answers with Claude~Opus~4.6 and retrieves with \texttt{text-embedding-3-large}, where \cognifold{} answers with \texttt{gpt-5.4-mini} and retrieves with the smaller \texttt{text-embedding-3-small}, holding extraction, embedding, and every step but the single answer to a small model. Its clearest deficit is temporal reasoning: at TR~$88.7$ it trails both systems above it ($95.5$ each) by nearly seven points, an event-time-arithmetic gap that is a write-path consolidation target rather than a read-path one. Absolute scores are not strictly comparable across rows---backbones span \texttt{gpt-4o-mini} to Claude~Opus~4.6, judges differ, and the vendor rows are self-reported---so the controlled axis remains the \emph{Build} model weighed against accuracy. The two together span both ends of conversational memory: a short coherent dialogue, and a long history thick with distractors.

\textit{General cognitive-domain probes.}
\begin{figure}[t]
  \centering
  \includegraphics[width=\linewidth]{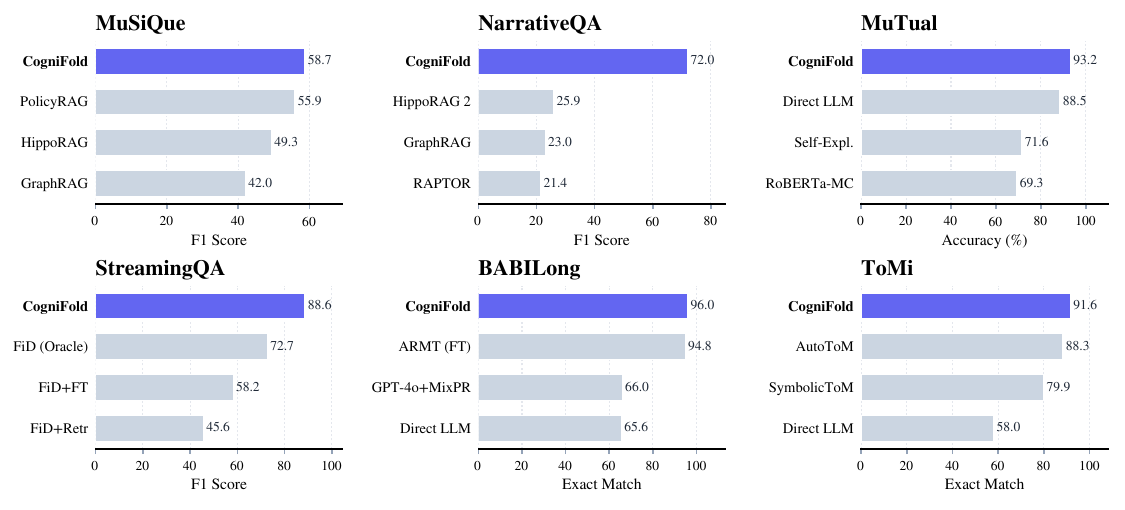}
  \caption{\textbf{Downstream benchmarks at a glance.} CogniFold (indigo, bold) against the most-cited published baselines for each benchmark, sorted by score with the best on top. Metric varies per benchmark; sample sizes are 500 for MuSiQue, NarrativeQA, MuTual, StreamingQA, ToMi, and 100 for BABILong.}
  \label{fig:benchmark_bars}
\end{figure}
Away from conversation, the graph faces the standard graph-retrieval suite of~\citet{gutierrez2025rag} on multi-hop reasoning and the most-cited published baselines elsewhere, all read with \texttt{gpt-4o-mini}. On MuSiQue it reaches F1~58.7, exceeding the strongest published RAG pipeline (HippoRAG\,2, +9.4) and the strongest symbolic-graph alternative (PolicyRAG, +2.8); on the others it leads on theory of mind (ToMi, +3.3 over AutoToM) and long-context factual extraction (BABILong, +1.2 over fine-tuned ARMT), holds within range of streaming-FiD on StreamingQA, and tops the published memory and structure-augmented baselines on MuTual and NarrativeQA. Figure~\ref{fig:benchmark_bars} summarises these six comparisons at a glance.

What the same substrate doing well across these very different cognitive domains shows is not a tuning result but a generality result. Conversational memory, multi-hop reasoning, theory of mind, narrative comprehension, and streaming temporal QA pull on different cognitive operations---inter-session consolidation, cross-document chaining, belief tracking, character disambiguation, time-anchored recall---yet they share the same underlying ask: that the memory substrate retain the right relational structure between events and surface it on demand. \cognifold{}'s consistent placement in the upper band across this spread is direct evidence that \emph{cognitive folding}---the operations of \S\ref{sec:debts}---is a task-general write-path competence rather than a benchmark-tuned heuristic. The one regime where the gain disappears is exactly the one the theory predicts: BABILong asks for verbatim local key--value supports, where structural folding adds no leverage that direct retrieval did not already have.

\section{Discussion}
\label{sec:discussion}

In a strict sense, \cognifold{} is beyond memory: it does not merely store and retrieve, it bootstraps. 
Each fold collapses events into higher-level concepts that later folds reason against; the agent is not starting cold at every event but using its own accumulated cognition as substrate. The 4.6$\times$ compression and 0.614 Proactivity (\S\ref{sec:cogeval_rq}) are the architectural correlates of this process, reflecting both the aggressive folding of events and how bootstrapped concepts further crystallize into goals.

\label{sec:limitations}
However, this bootstrapping dynamic introduces a principled limitation: because each fold conditions on previously accumulated structure, the same events in different orders produce different graphs. While this mirrors human \textbf{curriculum effects}---where pedagogical ordering yields cleaner schemas than shuffled inputs~\citep{elman1993learning, zhao2024model}---it raises an open question regarding memory stability. Order-aware consolidation, replay-based smoothing, and bounded-divergence analyses on streaming graphs are concrete avenues for future work.

A second limitation concerns the depth of our prefrontal mapping. \cognifold{} operationalizes only schema-driven integration, whereas the biological prefrontal cortex performs reward-based valuation, cognitive control, and counterfactual simulation~\citep{gilboa2017neurobiology, preston2013interplay}. Without value estimation, it cannot rank intents by long-horizon utility; without cognitive control, it cannot suppress impulsive emissions when a stronger goal is active; without counterfactual rollout, it cannot anticipate downstream consequences. Integrating these mechanisms forms the natural research arc beyond this paper.

\section{Conclusion}
\label{sec:conclusion}

We present \cognifold{}, an always-on proactive agent memory that folds fragmented events into persistent cognitive structure.
Unlike reactive retrieval systems, \cognifold{} builds a living graph that continuously folds, merges, decays, and reconnects under the event stream.
Because cognition grows recursively from the system's own products, goal-directed intents naturally emerge from converging evidence. We validated this design across two critical axes: proactive structural emergence on CogEval-Bench, and robust memory quality across eight broad-coverage downstream benchmarks.

As foundation-model capability grows, what a system computes in a single forward pass approaches what a human can reason about in a moment; what it accumulates, organizes, and bootstraps across time is where the value of an always-on agent will increasingly accrue. We release \cognifold{} as a foundation for research on real-time interaction, proactive collaboration, and agent cognition that bootstraps beyond memory.

\clearpage


\bibliographystyle{plainnat}
\bibliography{references}

\clearpage
\appendix

\section{Related Work}
\label{app:related}

\subsection{Graph-based Agent Memory}
\label{sec:rel_batch}
Current graph-augmented memory systems generally evolve along three axes, all of which treat the graph as a query-time artifact rather than a living substrate. HippoRAG~\citep{gutierrez2024hipporag, gutierrez2025rag} maps memory to a static knowledge graph retrieved via Personalized PageRank. However, this structurally frozen approach struggles to handle the evolving lifecycle of agent interactions.  For dynamic updates, Mem0~\citep{chhikara2025mem0} performs per-turn memory management through LLM-driven text rewrites. MAGMA~\citep{jiang2026magma} decomposes memory into multiple orthogonal graphs and supplements them with an LLM-driven slow-path inference pass. Zep~\citep{rasmussen2025zep} introduces bi-temporal validity to track fact invalidation. Across these paradigms, although several systems do process events incrementally, the graph's \textbf{topology} grows by accumulation, rewriting, or invalidation, but does not \textbf{metabolise} between events---it does not fold redundant fragments, decay stale connections by mere passage of time, or reconnect orphans through associative similarity. Real-time topological metabolism under an event stream therefore remains largely unaddressed.
Table~\ref{tab:paradigm} situates representative systems---together with the non-graph alternatives discussed below---on four orthogonal axes (stream input, proactive update, evolving topology, symbolic and inspectable), making the absent corner explicit.

\begin{table}[h]
\centering
\small
\setlength{\tabcolsep}{4pt}
\caption{\textbf{Design-space for agent memory under continuous streams.}
Four orthogonal axes distinguish always-on proactive memory from reactive, batch-oriented architectures; \cognifold{} is the only system satisfying all four simultaneously.
Table~\ref{tab:attack_surface} provides the complementary engineering-coverage view (which of the four structural debts of \S\ref{sec:debts} each system addresses).
\checkmark\,=\,satisfies; ---\,=\,does not; \emph{partial}\,=\,subset of the axis.}
\label{tab:paradigm}
\begin{tabular}{@{}lcccc@{}}
\toprule
System & Stream input & Proactive update & Evolving topology & Symbolic \\
\midrule
HippoRAG~\citep{gutierrez2024hipporag,gutierrez2025rag} & --- & --- & --- & \checkmark \\
Mem0~\citep{chhikara2025mem0} & partial & partial & --- & \checkmark \\
MAGMA~\citep{jiang2026magma} & --- & --- & partial & \checkmark \\
Zep~\citep{rasmussen2025zep} & \checkmark & \checkmark & partial & \checkmark \\
A-Mem~\citep{xu2025mem} & \checkmark & \checkmark & --- & \checkmark \\
Titans~\citep{behrouz2024titans} & \checkmark & \checkmark & n/a & --- \\
\midrule
\textbf{\cognifold{}} & \checkmark & \checkmark & \checkmark & \checkmark \\
\bottomrule
\end{tabular}
\end{table}


\subsection{Non-Topological Test-Time Learning}
\label{sec:rel_implicit}
A parallel line of work pursues persistent test-time learning through non-graph mechanisms, revealing a trade-off between inspectability and structural depth. Generative Agents~\citep{park2023generative} and MemGPT~\citep{packer2023memgpt} treat memory as natural language to be paged or searched, while systems like A-Mem~\citep{xu2025mem} rewrite Zettelkasten-style notes upon receiving new information. These text-rewriting methods remain highly inspectable but are structurally blind, as the underlying architecture does not evolve. Conversely, Titans~\citep{behrouz2024titans} introduces a neural long-term memory module updated via surprise-driven gradient descent during inference. While computationally efficient, this implicit approach deposits knowledge into an opaque weight space that cannot be audited or selectively deleted~\citep{mccloskey1989catastrophic}. Consequently, the field remains divided between methods that are transparent but topologically static, and those that learn continuously but sacrifice discrete, mutable geometry.

\subsection{Proactive Agent Memory}
\label{sec:rel_proactive}
Most agent-memory systems are fundamentally designed for the reactive setting: the user issues a query, and the system retrieves context to generate a response. Explicit treatment of the always-on setting---where input is a continuously arriving event stream---remains sparse. Traditional BDI-style architectures~\citep{bratman1987intention} address agency by deriving intentions from hardcoded rules and explicit goals, rather than letting them emerge bottom-up from accumulated evidence. Recent prospective-memory work~\citep{einstein2005prospective} and cognitive-agent frameworks~\citep{sumers2023cognitive, shinn2023reflexion} acknowledge the need for goal-directed organization but rarely anchor intent emergence to the evolving topology of the memory itself. Even with advanced read-path innovations, mainstream systems assume a stable graph at query time without defining how the structure should reorganize between queries, leaving a critical gap in anticipatory, autonomous memory assembly.

\section{Hyperparameters}
\label{app:hyperparams}

All numeric defaults referenced in \S\ref{sec:write_path} are listed here for reproducibility. Values were tuned empirically on a held-out subset of the personal-timeline simulator and held fixed across all reported experiments.

\subsection{Default Edge Weights (Table~\ref{tab:edge_types})}

\begin{table}[h]
\centering
\small
\begin{tabular}{@{}lc@{}}
\toprule
Edge Type & Default Weight \\
\midrule
\textsc{Grounds}, \textsc{Causes} & 0.9 \\
\textsc{Triggers}, \textsc{User\_Feedback} & 0.8 \\
\textsc{Reinforces}, \textsc{Part\_Of} & 0.7 \\
\textsc{Derived\_From}, \textsc{Deadline\_For} & 0.6 \\
\textsc{Related\_To} & 0.5 \\
\bottomrule
\end{tabular}
\caption{Prior weights at edge creation. Per-instance weights are then updated dynamically through \textsc{Reinforces} strengthening and exponential edge decay (\S\ref{sec:debts}).}
\label{tab:hyper_edges}
\end{table}

\subsection{Write-Path Scoring (Eq.~\ref{eq:score})}

\begin{table}[h]
\centering
\small
\setlength{\tabcolsep}{4pt}
\begin{tabular}{@{}lcc@{}}
\toprule
Parameter & Default & Description \\
\midrule
$\alpha$ (PageRank) & 0.4 & Structural importance \\
$\beta$ (Recency) & 0.4 & Temporal relevance \\
$\gamma$ (Access) & 0.2 & Usage frequency \\
$\lambda$ (Node decay) & 0.01\,/\,hr & Half-life $\approx$ 69 hours \\
$U$ (Urgency boost) & $[1.0, 2.0]$ & Linear ramp 24\,hr before deadline \\
$k$ (Context window) & 50 & Top-$k$ nodes selected for context \\
Tier proportions & 10\,/\,30\,/\,50\% & Immediate / Working / Background \\
Title-similarity dedup & 0.85 & Near-duplicate concept-merge threshold \\
\bottomrule
\end{tabular}
\caption{Write-path scoring weights and context-window allocation. Tier sub-weights: \emph{Immediate}---70\% recency + 30\% urgency; \emph{Working}---50\% PageRank + 30\% recency + 20\% type; \emph{Background}---80\% PageRank + 20\% diversity.}
\label{tab:hyper_writepath}
\end{table}

\subsection{Consolidation Operations (\S\ref{sec:debts})}

\begin{table}[h]
\centering
\small
\begin{tabular}{@{}lcc@{}}
\toprule
Parameter & Default & Description \\
\midrule
Merge cosine threshold & 0.85 & Minimum semantic similarity for two concepts to be merged \\
$\lambda_e$ (Edge decay) & 0.005\,/\,hr & Half-life $\approx$ 139 hours \\
kNN $k$ (completion) & 5 & Neighbours considered when inferring missing \texttt{GROUNDS} edges \\
kNN cosine threshold & 0.3 & Minimum similarity for an inferred kNN edge \\
\bottomrule
\end{tabular}
\caption{Consolidation operations that discharge the four structural debts. Default weights for each typed edge are listed in Table~\ref{tab:edge_types}.}
\label{tab:hyper_consolidation}
\end{table}

\subsection{Intent Emergence (Eq.~\ref{eq:ema})}

\begin{table}[h]
\centering
\small
\begin{tabular}{@{}lcc@{}}
\toprule
Parameter & Default & Description \\
\midrule
$\alpha_{\text{ema}}$ & 0.3 & EMA blend weight for per-category threshold updates \\
$s_t$ (accept) & 1.5 & Score on \emph{accept} feedback \\
$s_t$ (modify) & 1.1 & Score on \emph{modify} feedback \\
$s_t$ (defer) & 0.8 & Score on \emph{defer} feedback \\
$s_t$ (reject) & 0.3 & Score on \emph{reject} feedback \\
\bottomrule
\end{tabular}
\caption{Per-category EMA loop for adaptive intent-emission threshold.}
\label{tab:hyper_intent}
\end{table}

\subsection{Read Path}
\label{app:query}
The read path provides graceful fallback across four retrieval backends: BM25, semantic search (FAISS ANN), hybrid via Reciprocal Rank Fusion, and agentic multi-round search.

Graph traversal employs BFS with per-hop decay blended with query-seeded Personalized PageRank (parameters in Appendix~\ref{app:hyperparams}).

A 3-layer fast-path pipeline handles deployment: Layer\,1 ingests without LLM ($<$30s / 1{,}200 events), Layer\,2 batches LLM enrichment, and Layer\,3 computes embeddings.

Cost: ${\sim}\$0.02$/event; no GPU.

\begin{table}[h]
\centering
\small
\begin{tabular}{@{}lcc@{}}
\toprule
Parameter & Default & Description \\
\midrule
RRF $k$ & 60 & Reciprocal Rank Fusion smoothing constant \\
BFS hop decay & 0.85 & Per-hop weight decay during graph traversal \\
BFS / PPR blend & 60\,/\,40\% & Mix of BFS traversal and Personalized PageRank \\
\bottomrule
\end{tabular}
\caption{Read-path retrieval and traversal parameters.}
\label{tab:hyper_readpath}
\end{table}

\section{Four-Debt Attack Surface}
\label{app:attack_surface}

\begin{table}[t]
\centering
\small
\caption{\textbf{Four-debt attack-surface comparison.}
An event-stream memory graph accumulates four structural debts by the nature of its input (Section~\ref{sec:debts}):
\emph{accumulation}, \emph{compression}, \emph{decay}, \emph{completion}.
``---'' = not addressed; ``LLM rewrite'' = text-layer rewrite only; ``partial'' = subset of the debt; \checkmark = automatic graph-level operation.
\textbf{Scope of comparison}: all systems with a published description of their ingestion and update mechanisms in the 2024--2026 graph-memory literature; closed industrial systems without published mechanisms are excluded.
\cognifold{} is the only system to address all four debts as automatic graph-level operations.}
\label{tab:attack_surface}
\begin{tabular}{@{}lcccc@{}}
\toprule
System & Accumulate & Compress & Decay & Complete \\
\midrule
HippoRAG~\citep{gutierrez2024hipporag,gutierrez2025rag} & --- & --- & --- & partial (synonyms) \\
Cognee~\citep{cognee2024} & --- & --- & --- & --- \\
Mem0~\citep{chhikara2025mem0} & LLM rewrite & partial (write-time) & --- & --- \\
MAGMA~\citep{jiang2026magma} & --- & --- & --- & \checkmark\,(slow-path) \\
Zep / Graphiti~\citep{rasmussen2025zep} & --- & partial (summary) & ---\textsuperscript{$\dagger$} & --- \\
A-Mem~\citep{xu2025mem} & LLM rewrite & LLM rewrite & --- & partial (kNN+LLM) \\
PREMem & --- & LLM rewrite + dedup & --- & --- (no graph) \\
\midrule
\textbf{\cognifold{}} & \checkmark\,(\texttt{REINFORCES}) & \checkmark\,(\texttt{MERGE\_NODES}) & \checkmark\,(edge decay) & \checkmark\,(kNN inference) \\
\bottomrule
\end{tabular}
\begin{flushleft}\footnotesize
\textsuperscript{$\dagger$}\,Zep/Graphiti provides bi-temporal fact invalidation (\texttt{t\_invalid}) on \emph{contradiction}, which addresses \emph{consistency} rather than \emph{decay}: no edge weakens with passage of time alone. Consistency is orthogonal to the four debts in our taxonomy.
\end{flushleft}
\end{table}

Table~\ref{tab:attack_surface} compares how each prior agent-memory system addresses (or fails to address) the four structural debts of \S\ref{sec:debts}---accumulation, compression, decay, and completion. \cognifold{} is the first system to address all four as automatic, topology-level operations.

\section{Implementation}
\label{app:implementation}

Python 3.11: NetworkX (graph), Pydantic v2 (schemas), LangGraph (agent), FastAPI (HTTP), FAISS (ANN index). 21 packages, 1,167 tests, strict typing. Deployed on GCP Cloud Run with per-session isolation. Multi-domain support (learning, finance, programming) via prompt profiles.

\section{Core System Prompt (Excerpt)}
\label{app:prompt}

The LLM agent receives a system prompt composed from modular sections. The key section governing concept extraction:

\begin{lstlisting}[language={}, basicstyle=\ttfamily\scriptsize]
You are a cognitive graph agent. Given an event
and context, produce an UpdatePlan with operations:

NODE TYPES: event (raw input), concept (patterns),
  intent (goals), time (deadlines)
EDGE TYPES with weights:
  GROUNDS (0.9): event -> concept/intent
  REINFORCES (0.7): event -> existing concept
  TRIGGERS (0.8): concept -> intent
  PART_OF (0.7): concept -> concept (hierarchy)

RULES:
1. Create concepts for recurring patterns (3+ events)
2. Link every concept to grounding events
3. Merge near-duplicate concepts via MERGE_NODES
4. Create intents only when patterns suggest
   unmet goals with supporting evidence
5. Self-review: check for missing edges between
   concepts that share grounding events
\end{lstlisting}

The full prompt includes 20 composable sections (edge types, connectivity rules, validation checklist, deduplication, self-review). Domain-specific YAML profiles override the role section while retaining structural sections.

\section{CogEval-Bench Details}
\label{app:cogeval_details}

\subsection{Gold Graph Schemas}

For each scenario we manually define a gold concept graph $\mathcal{G}^* = (\mathcal{C}^*, \mathcal{R}^*, \mathcal{H}^*, \mathcal{I}^*)$ with four components.
\textbf{Concepts} $\mathcal{C}^*$: 8--9 non-hierarchical concepts, each specified with label, natural-language description, representative keywords, and expected event count.
Concepts are grounded in established domain knowledge (e.g., SoftEng includes Code Review, Sprint Planning, Deployment) and deliberately avoid \cognifold{}-specific abstractions to keep the gold standard system-agnostic.
\textbf{Relationships} $\mathcal{R}^*$: 9--14 labelled inter-concept edges using four relationship types drawn from cognitive schema theory~\citep{gilboa2017neurobiology, bartlett1995remembering}: \texttt{PART\_OF} (compositional hierarchy), \texttt{TRIGGERS} (temporal causation), \texttt{REINFORCES} (feedback strengthening), \texttt{CAUSES} (interventional causation).
\textbf{Hierarchy parents} $\mathcal{H}^*$: 1--3 superordinate concepts (e.g., Work Projects subsumes Coding Sessions and Code Review) that test hierarchical abstraction.
\textbf{Expected intents} $\mathcal{I}^*$: 2 goal nodes per scenario, each grounded in 2--3 supporting concepts with a specified trigger pattern (e.g., ``3+ exercise events within the week'' $\to$ Maintain Regular Exercise).
Each gold graph additionally contains 2 \textbf{planted multi-hop reasoning chains} (3--4 hops) following the compositional methodology of MuSiQue~\citep{trivedi2022musique}: chains are sequences of events connected through causal or temporal links across different concepts (e.g., deployment $\to$ staging bug $\to$ client demo failure), requiring cross-concept traversal to reconstruct.
Complete gold concept graphs for all 6 scenarios are provided in the supplementary material (\texttt{benchmarks/cogeval/data/gold\_graphs/}).

\subsection{Event Generation Pipeline}

From each gold graph, a grounded event stream is generated in four stages.
(1)~\textbf{Concept-grounded events}: for each concept $c \in \mathcal{C}^*$, GPT-4o-mini generates $n_c$ realistic first-person events conditioned on the concept's label, description, and keywords; each event receives the by-construction label $\text{gold\_concept} = c$.
(2)~\textbf{Chain events}: for each planted chain, the pipeline generates sequential events following the chain's step descriptions, ensuring entity consistency across hops (e.g., the same ``auth service v2.3'' appears in the deployment, bug discovery, and client demo events).
(3)~\textbf{Distractor injection}: 10--15\% of events come from unrelated topics (weather, unrelated news, etc.), labelled $\text{gold\_concept} = \texttt{null}$.
(4)~\textbf{Temporal shuffling}: all events receive timestamps spanning the scenario's temporal window (5--60 days), then are sorted chronologically with chain events interleaved among non-chain events to prevent trivial sequential pattern matching.

\subsection{Scenarios}

\begin{table}[t]
\centering
\caption{\textbf{CogEval-Bench scenarios.} Six scenarios across four domains, with controlled gold concept graphs. Each scenario defines gold concepts, planted multi-hop chains (3--4 hops), expected intents, and distractor events (${\sim}15\%$). Events generated by GPT-4o-mini from gold graphs, temporally shuffled.}
\label{tab:cogeval_scenarios}
\small
\begin{tabular}{@{}llcccc@{}}
\toprule
Scenario & Domain & Events & Gold Concepts & Chains & Intents \\
\midrule
\textsc{SoftEng} & Professional/tech work & 41 & 8 & 2 (3-hop) & 2 \\
\textsc{Health} & Medical recovery & 39 & 8 & 2 (3--4-hop) & 2 \\
\textsc{Team} & Product launch & 38 & 8 & 2 (3--4-hop) & 2 \\
\textsc{News} & Breaking news stream & 42 & 8 & 2 (3--4-hop) & 2 \\
\textsc{Academic} & Graduate research lab & 47 & 9 & 2 (3--4-hop) & 2 \\
\textsc{Support} & Customer support center & 44 & 8 & 2 (3--4-hop) & 2 \\
\midrule
\textbf{Total} & 4 domains & 251 & 49 & 12 & 12 \\
\bottomrule
\end{tabular}
\end{table}

Six scenarios span four domains (Table~\ref{tab:cogeval_scenarios}): professional work (SoftEng), medical recovery (Health), team coordination (Team), breaking news (News), academic research (Academic), and customer support (Support).
These domains test distinct cognitive patterns: daily routines with gradual concept consolidation, crisis cascades requiring causal chain tracking, topic drift across independent threads, deep abstraction hierarchies, and repetitive pattern detection.
The benchmark totals 251 events, 49 gold concepts, 12 planted multi-hop chains, and 12 expected intents across the six scenarios.

\subsection{Evaluation Metrics}

Three tracks are computed per system and averaged across scenarios.
\textbf{Track~A: Concept Emergence}---(i) \textbf{Gold~F1}: precision and recall of system concepts against gold concepts via embedding-based soft matching (\texttt{text-embedding-3-small}, cosine $\geq 0.75$) with optimal one-to-one assignment via the Hungarian algorithm~\citep{kuhn1955hungarian}; (ii) \textbf{LLM Quality}: GPT-4o-mini judge rates meaningfulness, groundedness, and abstraction level, each 0--1, then averaged; (iii) \textbf{Harmony}: harmonic mean of Gold~F1 and LLM Quality; (iv) \textbf{Purity}: average pairwise embedding similarity among events grounding each concept.
\textbf{Track~B: Relationship Topology}---Chain Discovery Rate (fraction of planted chains recoverable via BFS between endpoints), Clustering Coefficient~\citep{newman2006modularity}, Modularity (Newman Q), Edge Type Entropy.
\textbf{Track~C: Compression \& Proactivity}---Compression Ratio (input events / output concepts), PageRank Gini, Proactivity (fraction of intents with $\geq 2$ grounding connections).

\subsection{Comparison Systems}

Seven systems under identical LLM and events.
\textbf{OpenIE KG}: HippoRAG-style triples, flat entity graph, no concept folding.
\textbf{Cognee}~\citep{cognee2024}: Extract--Cognify--Load pipeline building property graphs via LLM entity extraction and classification; no cross-event merging or temporal tracking.
\textbf{HippoRAG\,2}~\citep{gutierrez2025rag}: deeper passage integration, synonym expansion edges, PPR retrieval; still entity-level with no concept abstraction.
\textbf{GraphRAG}~\citep{edge2024local}: LLM entity/relation extraction, Leiden community detection, LLM summarization per community; batch, no temporal folding or intent emergence.
\textbf{Mem0}~\citep{chhikara2025mem0}: text-rewrite memory cells with vector retrieval; no native graph (we materialise each cell as a node and induce vector-similarity edges).
\textbf{Zep}~\citep{rasmussen2025zep}: temporal knowledge-graph memory with entity-centric edges and time stamps.
\textbf{\cognifold{}}: full merge-fold pipeline (events $\to$ concepts $\to$ intents), 9 typed edges, online event-by-event processing, temporal decay, lifecycle management.

\subsection{Per-Scenario Results}

\begin{table}[t]
\centering
\caption{\textbf{CogEval-Bench per-scenario Harmony scores.} Harmony = harmonic mean of gold F1 and LLM quality (Track~A). \cognifold{} consistently leads across all 6 scenarios despite their diversity. Mem0's flat memory store (no concept abstraction) yields zero across the board; entity-level systems (OpenIE KG, Cognee, HippoRAG\,2, Zep/Graphiti) cluster around 0.07--0.19; community-level GraphRAG reaches 0.23--0.39; \cognifold{}'s folding achieves 0.38--0.57.}
\label{tab:cogeval_per_scenario}
\small
\setlength{\tabcolsep}{4pt}
\begin{tabular}{@{}l ccccccc@{}}
\toprule
Scenario & OpenIE KG & Cognee & HippoRAG\,2 & GraphRAG & Mem0 & Zep & \textbf{\cognifold{}} \\
\midrule
\textsc{SoftEng}   & .107 & .095 & .076 & .312 & .000 & .103 & \textbf{.572} \\
\textsc{Health}    & .138 & .090 & .101 & .309 & .000 & .081 & \textbf{.529} \\
\textsc{Team}      & .157 & .086 & .107 & .226 & .000 & .181 & \textbf{.383} \\
\textsc{News}      & .127 & .089 & .069 & .376 & .000 & .116 & \textbf{.433} \\
\textsc{Academic}  & .144 & .106 & .111 & .325 & .000 & .190 & \textbf{.423} \\
\textsc{Support}   & .158 & .100 & .105 & .390 & .000 & .158 & \textbf{.515} \\
\midrule
\textbf{Average}   & .138 & .094 & .095 & .323 & .000 & .138 & \textbf{.476} \\
\bottomrule
\end{tabular}
\\[6pt]
{\scriptsize
\setlength{\tabcolsep}{2.5pt}
\begin{tabular}{@{}l ccccccc ccccccc@{}}
\toprule
& \multicolumn{7}{c}{Compression Ratio $\uparrow$} & \multicolumn{7}{c}{Clustering Coefficient $\uparrow$} \\
\cmidrule(lr){2-8} \cmidrule(lr){9-15}
Scenario & KG & Cog & HR2 & GR & M0 & Zp & \textbf{CF} & KG & Cog & HR2 & GR & M0 & Zp & \textbf{CF} \\
\midrule
\textsc{SoftEng}   & 0.3 & 0.4 & 0.2 & 1.1 & 1.0 & 0.4 & \textbf{3.4} & .013 & .202 & .721 & .000 & .400 & .679 & \textbf{.367} \\
\textsc{Health}    & 0.2 & 0.4 & 0.2 & 1.3 & 1.0 & 0.5 & \textbf{4.9} & .000 & .202 & .684 & .000 & .452 & .742 & \textbf{.132} \\
\textsc{Team}      & 0.3 & 0.5 & 0.2 & 1.2 & 1.0 & 0.8 & \textbf{5.4} & .018 & .230 & .730 & .011 & .351 & .666 & \textbf{.316} \\
\textsc{News}      & 0.2 & 0.4 & 0.1 & 1.1 & 1.0 & 0.4 & \textbf{4.2} & .000 & .176 & .736 & .000 & .372 & .773 & \textbf{.372} \\
\textsc{Academic}  & 0.2 & 0.5 & 0.2 & 1.1 & 1.0 & 0.4 & \textbf{4.3} & .000 & .193 & .706 & .000 & .371 & .638 & \textbf{.380} \\
\textsc{Support}   & 0.3 & 0.5 & 0.2 & 1.3 & 1.0 & 0.6 & \textbf{5.5} & .000 & .200 & .720 & .000 & .347 & .726 & \textbf{.392} \\
\midrule
\textbf{Average}   & 0.3 & 0.5 & 0.2 & 1.2 & 1.0 & 0.5 & \textbf{4.6} & .005 & .200 & .716 & .002 & .382 & .704 & \textbf{.327} \\
\bottomrule
\end{tabular}
}
\\[4pt]
{\scriptsize KG = OpenIE KG; Cog = Cognee; HR2 = HippoRAG\,2; GR = GraphRAG; M0 = Mem0; Zp = Zep/Graphiti; CF = \cognifold{}. Compression = events/concepts ($\times$). Clustering marked \textbf{bold} for \cognifold{} due to semantic meaningfulness; HippoRAG\,2's high clustering arises from synonym edges, Zep's from dense entity co-mention edges, and Mem0's from undifferentiated similarity-induced edges (see text).}
\end{table}

Table~\ref{tab:cogeval_per_scenario} provides per-scenario breakdowns on three key metrics.
\cognifold{}'s Harmony ranges from 0.383 (Team) to 0.572 (SoftEng), consistently above GraphRAG's best (0.390, Support).
Compression is robust (3.4--5.5$\times$), highest on Support and Team where repetitive event patterns benefit most from folding.
Clustering ranges from 0.132 (Health) to 0.392 (Support), lowest on Health where longer temporal spans create sparser event overlap.

\subsection{Why Enrichment $\neq$ Abstraction}

Cognee and HippoRAG\,2 occupy the same Harmony tier as vanilla OpenIE KG despite adding substantial computational machinery (ECL pipeline, synonym expansion, Personalized PageRank).
This confirms that the bottleneck in concept emergence is not extraction depth but \textbf{cross-event integration}---the ability to recognize that events $e_1, e_3, e_7$ ground the same underlying concept and merge them into a single abstraction.
Systems that process each event in isolation, however thoroughly, cannot produce this integration.

\subsection{Proactivity as Emergence Evidence}

Proactivity captures a property no baseline can produce: identifying goals from converging evidence before being asked.
An average of 22 intent nodes per scenario, with 61.4\% well-grounded by multiple events, constitutes evidence of intent emergence beyond memory organization.
While these intents are LLM-generated rather than autonomously emergent from prediction error, they represent a measurable step toward proactive intelligence that flat memory systems cannot take.

\subsection{Construct Validity}

Two metrics that most sharply separate \cognifold{} from baselines---Purity (event-level grounding) and Proactivity (intents with $\geq 2$ groundings)---measure representational features that only \cognifold{} emits: \texttt{GROUNDS} edges for Purity, intent nodes for Proactivity.
A reader may reasonably worry the metric is biased toward our representation.
We address this risk with three choices: (i) the representational hierarchy holds not only on these two metrics but also on task-agnostic ones (Harmony, Gold~F1, LLM Quality, Compression, Clustering), on which baselines are free to score well; (ii) CogEval is deliberately supplemented by the broad eight-benchmark memory results (\S\ref{sec:exp_downstream}), which use external benchmarks and metrics; (iii) we report raw intent counts and grounding densities, not normalized scores.
Even so, a fully adversarial benchmark---one where our representation could lose---is future work.

\subsection{Scale and Scope Limitations}

CogEval-Bench uses synthetic events generated from predefined gold graphs, which may not capture the full complexity of natural event streams.
The scale is deliberately small (${\sim}42$ events per scenario) to enable controlled structural evaluation; whether the structural hierarchy holds at larger scales remains to be validated.
The LLM judge introduces potential evaluation bias; we mitigate this by per-concept independent scoring (not pairwise comparison) and report the full judge prompt above.

\subsection{Event Generation Prompt}

For each concept $c$ with label $\ell$ and description $d$, events are generated using the following prompt template:

\begin{lstlisting}[language={}, basicstyle=\ttfamily\scriptsize]
Generate {n} realistic first-person events for a
{scenario} scenario, grounded in the concept
"{label}": {description}.

Each event should:
- Be a specific, timestamped experience (not generic)
- Use first-person perspective
- Be self-contained (understandable without context)
- Relate clearly to the concept keywords: {keywords}

Output as JSON array with fields: title, description,
timestamp, event_type.
\end{lstlisting}

Chain events use a sequential prompt that references entities from prior chain steps to maintain cross-hop consistency. Distractor events are generated from unrelated topics (e.g., weather, unrelated hobbies) without concept grounding.

\subsection{LLM Judge Prompt}
\label{app:judge_prompt}

The concept quality judge evaluates each system concept independently:

\begin{lstlisting}[language={}, basicstyle=\ttfamily\scriptsize]
Evaluate this concept extracted from an event stream.

Concept: "{concept_label}"
Grounding events (if any): {event_summaries}
Scenario context: {scenario_description}

Rate on three dimensions (0.0 to 1.0):

1. MEANINGFULNESS: Is this a semantically coherent
   concept? (1.0 = clearly defined theme/pattern,
   0.0 = incoherent or trivial)

2. GROUNDEDNESS: Is the concept well-supported by
   its grounding events? (1.0 = strong evidence,
   0.0 = no supporting evidence)

3. ABSTRACTION LEVEL: Is this the right level of
   abstraction? (1.0 = useful generalization,
   0.0 = too specific or too vague)

Output JSON: {"meaningfulness": X, "groundedness": X,
  "abstraction": X}
\end{lstlisting}

\subsection{Embedding Similarity Threshold Sensitivity}
\label{app:sensitivity}

The Gold F1 computation uses cosine similarity $\geq 0.75$ for concept matching. Table~\ref{tab:sensitivity} reports \cognifold{}'s Harmony score averaged across 6 scenarios at varying thresholds, confirming that the ranking is robust to threshold choice.

\begin{table}[h]
\centering
\small
\caption{Sensitivity of Harmony to embedding similarity threshold (CogniFold, averaged over 6 scenarios).}
\label{tab:sensitivity}
\begin{tabular}{@{}lccccc@{}}
\toprule
Threshold & 0.65 & 0.70 & \textbf{0.75} & 0.80 & 0.85 \\
\midrule
Harmony & 0.501 & 0.489 & \textbf{0.476} & 0.452 & 0.418 \\
\bottomrule
\end{tabular}
\end{table}

Higher thresholds are more conservative (fewer matches, lower recall); lower thresholds are more permissive (more matches, risk of false positives). The relative ranking of systems is preserved across all tested thresholds.

\section{Reproducibility}
\label{app:reproducibility}

\textbf{Code and data.} Source code, benchmark runner scripts, and evaluation harnesses will be released upon publication. All benchmarks use publicly available datasets: MuTual~\cite{cui2020mutual}, ToMi~\cite{le-etal-2019-revisiting}, MuSiQue~\cite{trivedi2022musique}, NarrativeQA~\cite{kovcisky2018narrativeqa}, StreamingQA~\cite{liska2022streamingqa}, LoCoMo~\cite{maharana2024evaluating}, LongMemEval~\cite{wu2025longmemeval}, and BABILong~\cite{kuratov2024babilong}.

\textbf{Compute.} All experiments use the OpenAI API (GPT-4o-mini for generation---except the LongMemEval answer path, which uses \texttt{gpt-5.4-mini}; memory construction stays GPT-4o-mini throughout---and text-embedding-3-small for embeddings). Graph construction and retrieval run on a single CPU; no GPU is required. Total API cost for all experiments (including development iterations): approximately \$150.

\textbf{Randomness.} The primary source of non-determinism is LLM sampling (temperature 0.0 for all evaluation calls). Graph construction is deterministic given the same LLM outputs. Benchmark sampling uses fixed random seeds (seed=42 for all dataset splits). Wilson confidence intervals are reported for full-scale results ($n{\geq}100$).

\textbf{Broader impact.} \cognifold{} is a general-purpose cognitive memory architecture. Like any persistent memory system, it raises privacy considerations: the graph accumulates personal information that persists across sessions. Production deployments should implement access controls and data retention policies. The system does not autonomously take actions; intent nodes represent identified goals, not executed plans.

\end{document}